\documentclass[10pt,twocolumn,letterpaper]{article}

\usepackage[pagenumbers]{cvpr} 

\usepackage{microtype}

\renewcommand{\paragraph}[1]{\vspace{.3em}\noindent\textbf{#1.}}

\newlength{\defaultabovecaptionskip}
\newlength{\defaultbelowcaptionskip}
\AtBeginDocument{
    \setlength{\defaultabovecaptionskip}{\abovecaptionskip}
    \setlength{\defaultbelowcaptionskip}{\belowcaptionskip}
}

\usepackage{xspace}
\newcommand{\paper}{TimeLens\xspace}

\usepackage{multirow}
\usepackage{graphicx}
\usepackage[table,xcdraw]{xcolor}
\usepackage{siunitx}
\usepackage[normalem]{ulem}
\useunder{\uline}{\ul}{}

\usepackage{marvosym}

\definecolor{Gray}{gray}{0.5}
\definecolor{LGray}{gray}{0.9}
\definecolor{darkblue}{RGB}{94,110,186}
\definecolor{darkGreen}{RGB}{92, 148, 110}
\definecolor{myblue}{RGB}{14, 121, 178}
\definecolor{rightpath}{RGB}{248, 203, 173}
\definecolor{wrongpath}{RGB}{89, 89, 89}

\usepackage[most]{tcolorbox}

\newcommand{\findingz}[2]{
    \vspace{0.2em}
    \begin{tcolorbox}[
        colback=blue!5,
        colframe=blue!60!black,
        arc=2pt,
        boxsep=2pt,
        left=5pt, right=5pt,
        top=2pt, bottom=2pt,
        boxrule=0.6pt,
        drop shadow={opacity=0.25},
        enhanced jigsaw
    ]
    \textbf{\textit{Finding #1:}} #2
    \end{tcolorbox}
    \vspace{0.15em}
}

\definecolor{cvprblue}{rgb}{0.21,0.49,0.74}
\usepackage[pagebackref,breaklinks,colorlinks,allcolors=cvprblue]{hyperref}

\def\paperID{} 
\def\confName{CVPR}
\def\confYear{2026}

\title{TimeLens: Rethinking Video Temporal Grounding with Multimodal LLMs}

\author{
\textbf{Jun Zhang}\textsuperscript{1,2,*} \quad
\textbf{Teng Wang}\textsuperscript{2,~\Letter} \quad
\textbf{Yuying Ge}\textsuperscript{2}  \quad
\textbf{Yixiao Ge}\textsuperscript{2} \quad
\textbf{Xinhao Li}\textsuperscript{1}
\\[0.15em]
\phantom{XX}
    \textbf{Ying Shan}\textsuperscript{2}
    \hspace{3em}  
    \textbf{Limin Wang}\textsuperscript{1,3,~\Letter}
\\[0.15em]
$^1$Nanjing University \quad
$^2$ARC Lab, Tencent PCG \quad
$^3$Shanghai AI Lab
\\[0.2em]
\url{https://timelens-arc-lab.github.io/}
}

\begin{document}
\maketitle
\begin{abstract}

This paper does not introduce a novel method but instead establishes a straightforward, incremental, yet essential baseline for video temporal grounding (VTG), a core capability in video understanding.
While multimodal large language models (MLLMs) excel at various video understanding tasks, the recipes for optimizing them for VTG remain under-explored.
In this paper, we present \textbf{TimeLens}, a systematic investigation into building MLLMs with strong VTG ability, along two primary dimensions: data quality and algorithmic design.
We first expose critical quality issues in existing VTG benchmarks and introduce \textbf{TimeLens-Bench}, comprising meticulously re-annotated versions of three popular benchmarks with strict quality criteria. Our analysis reveals dramatic model re-rankings compared to legacy benchmarks, confirming the unreliability of prior evaluation standards. We also address noisy training data through an automated re-annotation pipeline, yielding \textbf{TimeLens-100K}, a large-scale, high-quality training dataset.
Building on our data foundation, we conduct in-depth explorations of algorithmic design principles, yielding a series of meaningful insights and effective yet efficient practices. These include interleaved textual encoding for time representation, a thinking-free reinforcement learning with verifiable rewards (RLVR) approach as the training paradigm, and carefully designed recipes for RLVR training. These efforts culminate in \textbf{TimeLens models}, a family of MLLMs with state-of-the-art VTG performance among open-source models and even surpass proprietary models such as GPT-5 and Gemini-2.5-Flash.
All codes, data, and models will be released to facilitate future research.

\end{abstract}
{
\renewcommand{\thefootnote}%
{\fnsymbol{footnote}}
\footnotetext[0]{\Letter~Corresponding author. *~Work done during internship at ARC Lab, Tencent PCG.}
}
\setlength{\abovecaptionskip}{3pt}
\setlength{\belowcaptionskip}{1pt}
\vspace{-1em}
\section{Introduction}
\label{sec:intro}


Recent multimodal large language models (MLLMs) have excelled at understanding ``what" happens in a video, yet they largely fail when asked ``when." This limitation is central to the task of video temporal grounding (VTG).
The challenge is twofold: 1) VTG necessitates a fundamental shift from coarse semantic aggregation to fine-grained time-aware perception; 2) Distinguishing queried events requires modeling long-term visual dynamics over appearance-centric features, which are notoriously difficult to annotate and learn. As MLLMs become integral to perception~\cite{yuan2021closerlookat,wu2025numberit,perception_test,vcr_bench} and reasoning systems~\cite{nagrani2025minerva,nagrani2024neptune,liu2025videomind,chain_of_frames,zhang2025vitcot,longvila_rl}, equipping them with robust temporal awareness is no longer optional, but essential~\cite{vtg_survey,univtg,timechat,wang2025timer1,team2025vidi}. 

\begin{figure}[t]
  \vspace{1em}
  \centering
   \includegraphics[width=1.0\linewidth]{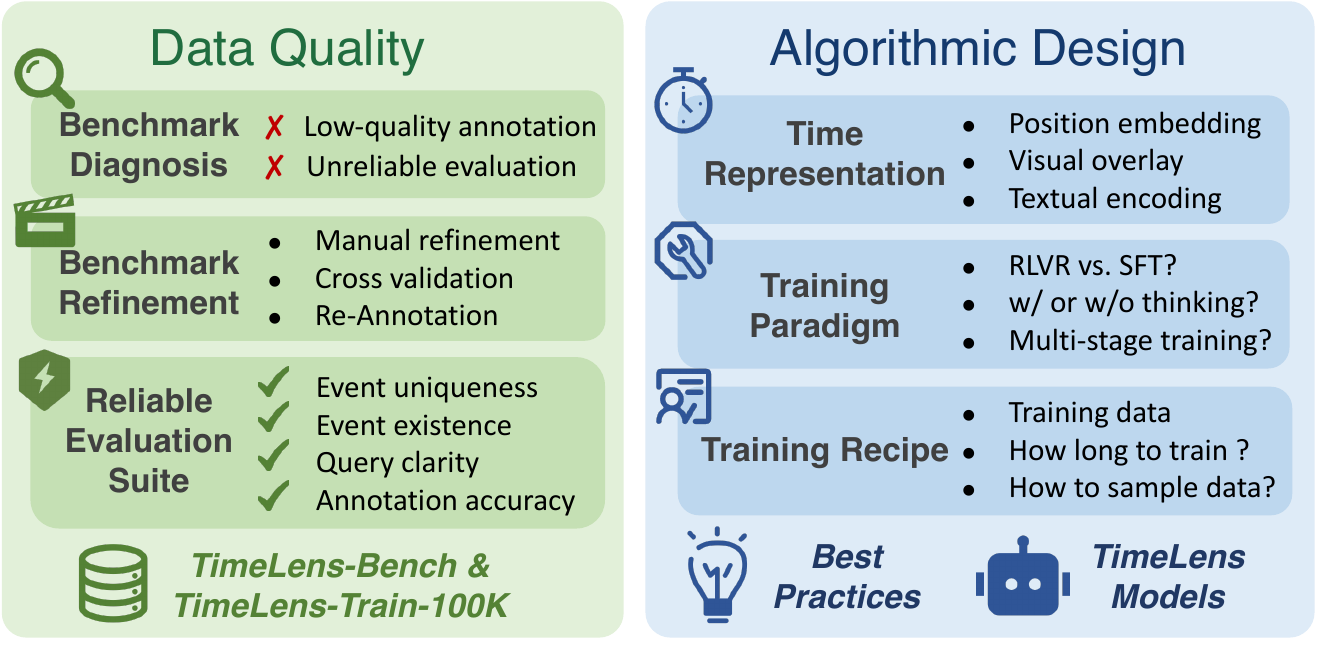}
   \caption{\textbf{Overview of the proposed TimeLens framework}. We systematically explore the key factors for building performant video temporal grounding models, dissecting our efforts along two primary dimensions: data quality and algorithmic design. 
   For data quality, we focus on benchmark diagnosis, benchmark refinement, and creating a reliable evaluation suite. For algorithmic design, we study various aspects including time encoding, training recipes, and optimization strategies to establish best practices and develop the TimeLens models.
   }
   \label{fig:teaser1}
   \vspace{-1em}
\end{figure}

\begin{figure*}[t]
  \centering
  \begin{subfigure}[b]{0.48\linewidth}
     \centering
     \includegraphics[width=\textwidth]{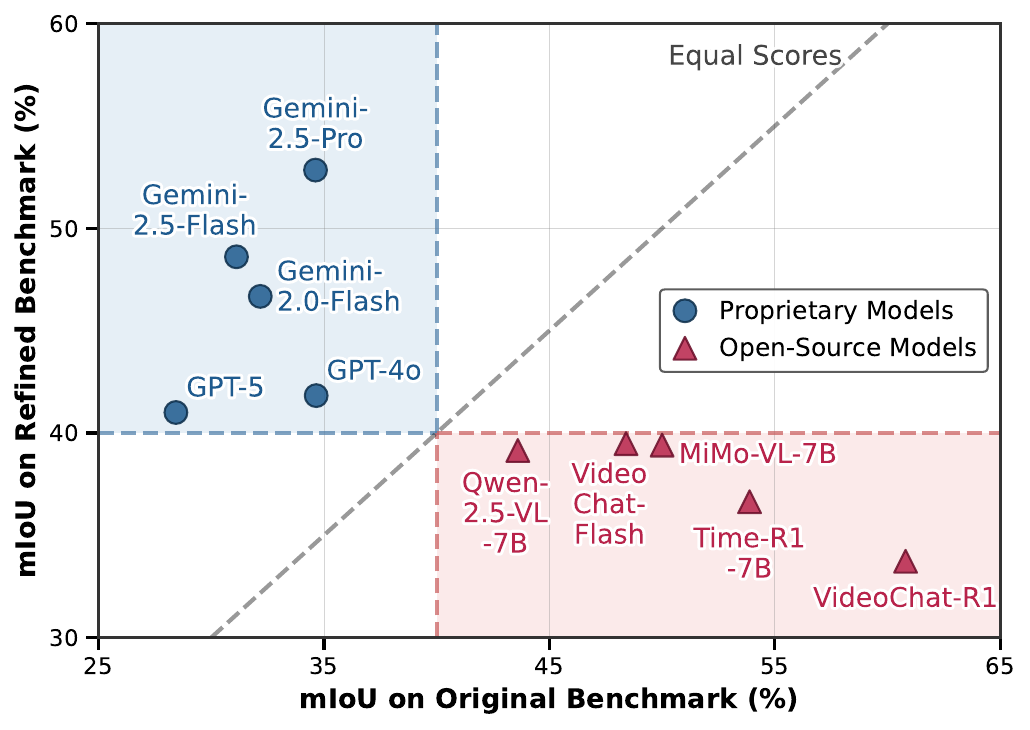}
     \caption{\ }
     \label{fig:performance_charades}
  \end{subfigure}
  \hfill 
  \begin{subfigure}[b]{0.51\linewidth}
     \centering
     \includegraphics[width=\textwidth]{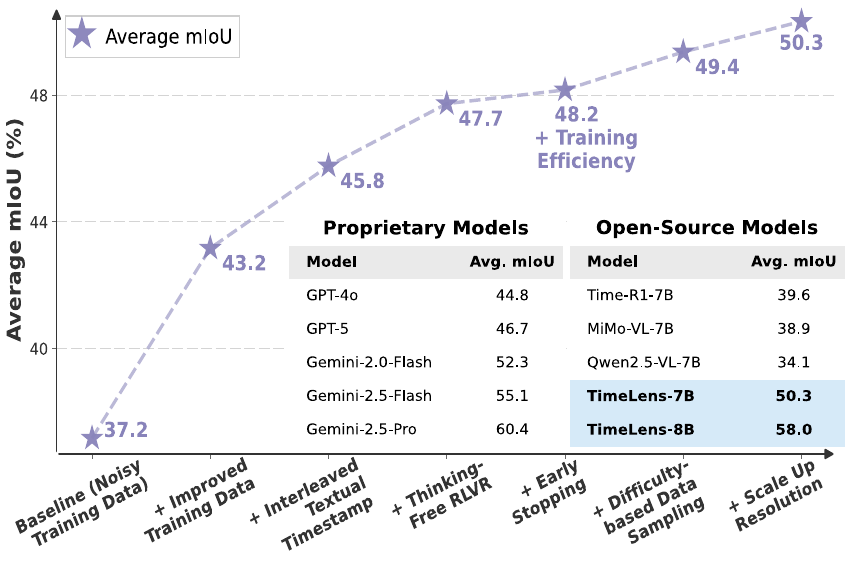}
     \caption{\ }
     \label{fig:abalation_adding}
  \end{subfigure}
  \label{fig:two_subs}

  \caption{
    \textbf{(a) Impact of data quality on model evaluation.} A comparison of Mean IoU on original versus our refined Charades-STA benchmarks. The deviation from the diagonal line shows that legacy benchmarks are misleading, as they inflate the results of some open-source models while underestimating proprietary ones. 
    \textbf{(b) Cumulative performance gains of TimeLens explorations.} This analysis shows how each component boosts the model's average performance on TimeLens-Bench. From data curation to thinking-free RLVR with early stopping and difficulty-based data sampling, each step demonstrates a clear positive impact towards our final TimeLens model. TimeLens-7B and TimeLens-8B are based on Qwen2.5-VL-7B and Qwen3-VL-8B, respectively.
}
\end{figure*}

This work focuses on post-training MLLMs with leading temporal grounding ability. This investigation is a straightforward extension given the recent progress in pretrained foundation MLLMs~\cite{qwen2-5-vl,Qwen3-VL,internvideo2}. 
Different from heavily studied general understanding tasks, recipes for fine-grained grounding tasks are not yet to be established. This paper aims to systematically investigate core components of building time-aware MLLMs~(Fig.~\ref{fig:teaser1}) along two primary dimensions: \textbf{data quality} and \textbf{algorithmic design}.

Our investigation starts by exposing critical flaws in evaluation benchmarks. We find that existing VTG benchmarks~\cite{charades-sta,activitynet-captions,qvhighlights} not only lack a clear comparison between leading proprietary and open-source models but are also rife with low-quality queries and erroneous timestamps. This noisy data may render current leaderboards misleading and misguide research efforts. To rectify this, we undertook a meticulous data overhaul. We first defined strict criteria for query and timestamp quality, in terms of uniqueness, existence, clarity, and accuracy. We then manually re-annotated three popular datasets (Charades-STA~\cite{charades-sta}, ActivityNet Captions~\cite{activitynet-captions}, QVHighlights~\cite{qvhighlights}) to create \textbf{TimeLens-Bench}, a rigorously cross-validated benchmark. 
%
As shown in Fig.~\ref{fig:performance_charades},
the necessity of this correction is confirmed by a dramatic re-ranking of models on TimeLens-Bench compared to their performance on legacy benchmarks, proving the unreliability of prior evaluation standards.
Beyond evaluation, we also fix the noisy training data by automated re-annotation, yielding \textbf{TimeLens-100K}, a large-scale, high-quality training dataset.

With our curated data suite as a solid foundation, we conduct in-depth explorations on the algorithmic design principles from three key aspects. 
First, for timestamp representation, we discover that a simple yet effective interleaved textual encoding strategy outperforms more complex alternatives. 
Second, we determine that VTG is fundamentally a perception-driven task, and thus employ a pure thinking-free reinforcement learning with verifiable rewards (RLVR) approach that outperforms other training paradigms in both efficiency and performance. 
Finally, our detailed analysis of RLVR training reveals two key recipes for both performance and training efficiency: (1) early stopping when reward metrics plateau, and (2) difficulty-based data sampling. By integrating these insights and design principles, we ultimately develop \textbf{\paper models}, a family of MLLMs with superior VTG capability. As shown in \cref{fig:abalation_adding}, our model achieves state-of-the-art performance among open-source models and even surpasses proprietary models such as GPT-5 and Gemini-2.5-Flash.

Through these efforts, we identified and addressed long-overlooked quality issues in existing datasets, and derived a series of insights and best practices in algorithmic design. We hope TimeLens can serve as a solid foundation in both data curation and algorithmic design principles, to facilitate future research on building MLLMs with strong VTG capabilities. Our code, data, and models will be open-sourced.

\begin{figure*}[t]
  \centering
   \includegraphics[width=\linewidth]{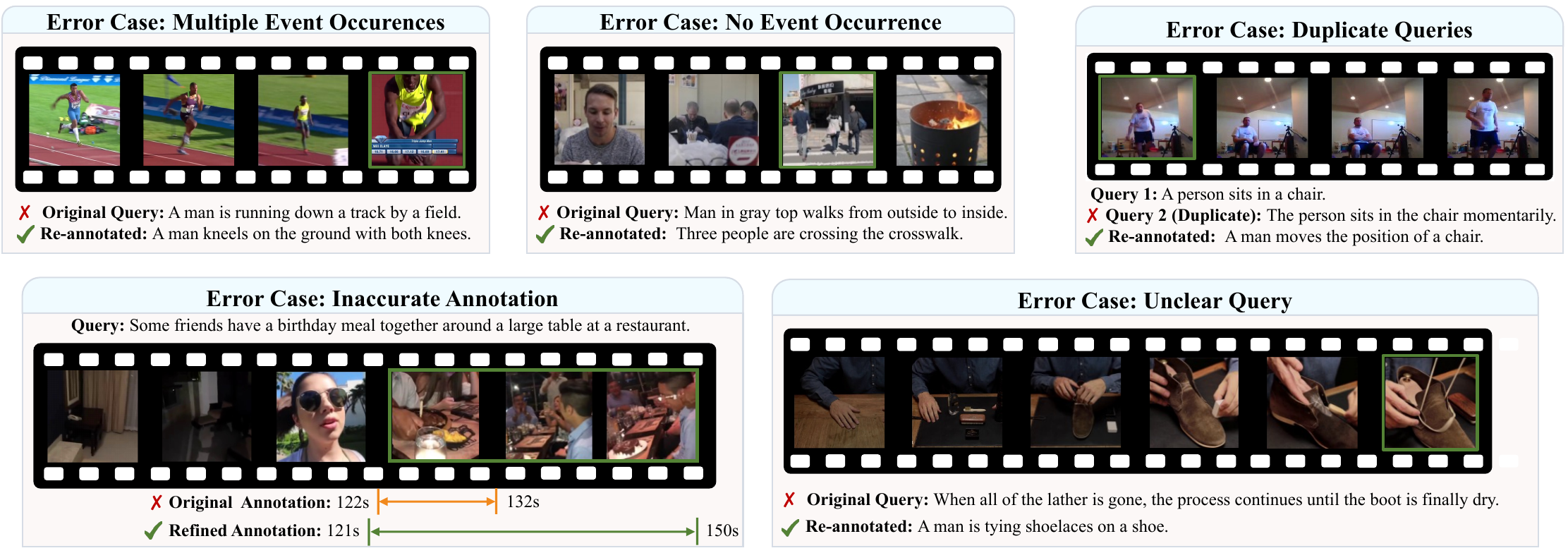}
    \caption{\textbf{Qualitative examples of errors and fixes.} We present representative errors identified in existing datasets, spanning different error types, including multiple event occurrences, no event occurrence, duplicate queries for the same video, unclear query, and inaccurate annotation. Through our rigorous manual refinement, these errors have been properly corrected, significantly improving data quality.}
  \label{fig:error_qualitative}
\end{figure*}


\section{Related Work}

\paragraph{Temporal Grounding Datasets}
Numerous VTG datasets have been proposed, spanning diverse domains~\cite{qvhighlights,activitynet-captions,oncescu2021queryd,wang2024cosmo_cap,huang2024vtimellm-internvid-vtime,grauman2022ego4d,tacos}.
Early works~\cite{charades-sta,2d-tan,lu2019debug} trained and evaluated models on the training and test splits of a single benchmark~\cite{tacos,activitynet-captions} to assess their ability to fit single-domain data distribution. In recent works~\cite{liu2025videomind,timechat,guo2024trace}, large diverse corpuses composed of multiple different source datasets ~\cite{didemo,oncescu2021queryd,wang2024cosmo_cap,huang2024vtimellm-internvid-vtime,zala2023hirest,liu2024etbench} are aggregated for training, and a suite of distinct benchmarks~\cite{activitynet-captions,charades-sta,qvhighlights} are used to probe the models' real-world cross-domain generalizability.

However, the critical issue of data quality has been overlooked. There lacks a systematic examination on whether existing datasets are reliable enough for training and evaluation. In this paper, we manually inspect existing datasets, identify and correct errors, and produce quality-improved training and evaluation suites for developing more practical VTG models.

\paragraph{MLLMs for Temporal Grounding}
Substantial works focus on algorithmic designs to improve MLLMs' VTG capability. One line of research explores model architectures, including token compression methods to reduce computation on long videos ~\cite{timechat,zeng2024timesuite}, timestamp encoding strategies to align the timestamps of each frame with its corresponding features~\cite{chen2024timemarker,wu2025numberit,li2025unitime,ge2025arc-hunyuan-video,zeng2025distime,wang2024grounded_videollm,li2025llava_st}. Another line of works investigate training strategies: introducing various supervised fine-tuning tasks to improve VTG performance~\cite{zeng2024timesuite,cheng2025tempura}, or designing verifiable rewards to improve performance via reinforcement learning~\cite{wang2025timer1,tvg-r1,li2025tempsamp_r1,yue2025tempo_r0}.

Despite the abundance of proposed designs, their inconsistent experimental settings make it difficult to fairly compare their relative merits and establish best practices. In this paper, we systematically analyze these design choices using our quality-assured training and evaluation suites, offering key insights for improving MLLMs’ VTG capability.

\begin{figure*}[t]
  \centering
   \includegraphics[width=1\linewidth]{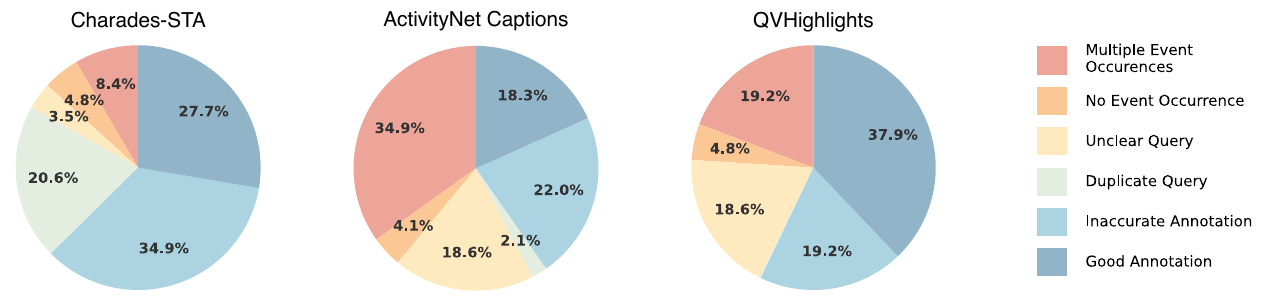}
   \caption{\textbf{Statistics of errors indicating alarmingly high proportion of errors in existing datasets.}
   }
   \label{fig:error_analysis}
\end{figure*}

\section{Towards Reliable, High-Quality VTG Data}

\subsection{Annotation Criteria}
\label{sec:criteria}

\paragraph{Task Formulation} For temporal grounding, a model takes as input a video~$v$ and a text query~$q$, localizes the event~$E$ described by~$q$, and outputs the corresponding temporal segment $S = (t_{\text{start}}, t_{\text{end}})$. In practice, a video is typically annotated with one or more query-segment pairs~$\{(q_i, S_i)\}_{i=1}^n$.

\paragraph{Input Criteria}
The input video and query should satisfy:
\begin{itemize}
    \item {
        \textit{{Query clarity and specificity}}. The query must be clear, precise, and unambiguous for accurate and definitive grounding (A counterexample like ``the game continues'').
    }
    \item{
        \textit{Event existence}. The event described in the text query must genuinely exist within the video content.
    }
    \item{
        \textit{Query uniqueness}. All queries must be unique in a single video. The presence of multiple nearly identical queries describing the same event is equivalent to duplicating or weighting certain samples, leading to biased metrics.
        Indeed, this issue is severe in Charades-STA dataset.
    }
    \item {
        \textit{Avoid information leakage in queries}. Queries like ``ending credits'' leak their temporal position, allowing the model to answer via shortcut, without truly ``grounding'' the query over the entire video. However, annotators tend to label such queries since they are easy to identify.
    }
\end{itemize}

\paragraph{Output Criteria}
The temporal segment should satisfy:
\begin{itemize}
    \item {
        \textit{Annotation precision}. The annotated event boundaries should be precise, excluding any subsegments that do not conform to the query's description.
    }
    \item {
        \textit{Annotation exhaustiveness}. There should be no other time segments outside the annotated one that also satisfy the query's description.
    }
\end{itemize}

\subsection{Manual Auditing and Refinement}
\label{sec:manual_pipeline}

We introduce a rigorous and efficient pipeline for auditing and refining existing temporal grounding datasets.

\paragraph{Diagnose-then-Refine}
Our pipeline follows a \textit{diagnose-then-refine} workflow. Given a video-query pair from existing datasets, annotators first carefully review the video to identify potential errors against the criteria in~\cref{sec:criteria}. If an error is detected, they select the error category, then either revise the query or choose a new valid event to describe. Subsequently, the precise temporal segment is annotated.
The core principle is that the \textbf{same} annotator performs both error detection and subsequent correction, which not only improves efficiency but also strengthens annotators' awareness of potential errors, thereby reducing the risk of introducing similar ones.

\paragraph{Error Identification}
Directly applying the abstract criteria from~\cref{sec:criteria} for error detection proves overly challenging for annotators. Therefore, as shown in \cref{fig:error_qualitative}, we derive from these criteria a set of concrete, easily identifiable error types with clear illustrations. Annotators check whether each error type is present and fill in the corresponding information. Additionally, we group all queries from the same video together to detect violations of ``query uniqueness'' and improve annotation efficiency. During the process, we do not provide original temporal segments to annotators.

\paragraph{Quality Control}
Upon completion of each small data batch, every sample is assigned to a different annotator for cross-validation and error correction. If the error rate in a batch exceeds a threshold, the entire batch is rejected for re-annotation and then validated again.
%
For annotator selection and training, we sampled a small subset of data for trial annotation with over a dozen vendors, then selected the vendor with the highest quality and consistency. Before formal annotation, we provided a detailed handbook and conducted several training sessions.
The annotation interface and detailed manual are provided in \cref{seg:anno_tool} of the appendix.


\begin{table*}[h]
\centering
\small
\setlength{\tabcolsep}{2.4pt}
\renewcommand{\arraystretch}{1.0} 
\begin{tabular}{lcccc|cccc|cccc}
\toprule
                                                & \multicolumn{4}{c|}{\textbf{Charades-TimeLens}}                                            & \multicolumn{4}{c|}{\textbf{ActivityNet-TimeLens}}                                         & \multicolumn{4}{c}{\textbf{QVHighlights-TimeLens}}                                        \\ \cmidrule(l){2-13}
\multirow{-2}{*}{\textbf{Model}}                & R1@0.3               & R1@0.5               & R1@0.7               & mIoU                  & R1@0.3               & R1@0.5               & R1@0.7               & mIoU                  & R1@0.3               & R1@0.5               & R1@0.7               & mIoU                 \\ \midrule
\textit{Proprietary Models}                     &                      &                      &                      &                       &                      &                      &                      &                       &                      &                      &                      &                      \\
GPT-4o~\cite{hurst2024gpt-4o}                   & 60.6                 & 44.5                 & 23.5                 & 41.8                  & 55.2                 & 41.4                 & 25.8                 & 40.4                  & 69.0                 & 54.8                 & 38.5                 & 52.1                 \\
GPT-5~\cite{OpenAI2025_GPT5}                    & 59.3                 & 42.0                 & 22.0                 & 40.5                  & 57.4                 & 44.9                 & 30.4                 & 42.9                  & 72.4                 & 60.4                 & 46.4                 & 56.8                 \\
Gemini-2.0-Flash~\cite{comanici2025gemini-2-5}  & 66.4                 & 53.5                 & 27.1                 & 46.7                  & 62.9                 & 54.0                 & 37.7                 & 49.3                  & 76.2                 & 66.4                 & 48.3                 & 60.8                 \\
Gemini-2.5-Flash~\cite{comanici2025gemini-2-5}  & 68.7                 & 56.1                 & 30.6                 & 48.6                  & 66.8                 & 57.5                 & 41.3                 & 52.5                  & 78.2                 & 69.4                 & 55.0                 & 64.3                 \\
Gemini-2.5-Pro~\cite{comanici2025gemini-2-5}    & 74.1                 & 61.1                 & 34.0                 & 52.8                  & 72.3                 & 64.2                 & 47.1                 & 58.1                  & 84.1                 & 75.9                 & 61.1                 & 70.4                 \\ \midrule
\textit{Open-Source Models}                     &                      &                      &                      &                       &                      &                      &                      &                       &                      &                      &                      &                      \\
VideoChat-Flash-7B~\cite{li2024videochat_flash} & 60.2                 & 37.9                 & 17.8                 & 39.7                  & 35.5                 & 21.8                 & 10.5                 & 24.8                  & 45.2                 & 30.6                 & 16.7                 & 32.7                 \\
VideoChat-R1-7B~\cite{li2025videochat_r1}       & 51.9                 & 30.8                 & 11.7                 & 33.7                  & 35.0                 & 23.9                 & 11.3                 & 25.0                  & 29.3                 & 19.1                 & 9.4                  & 21.5                 \\
Time-R1-7B~\cite{wang2025timer1}                & 57.9                 & 32.0                 & 16.9                 & 36.6                  & 44.8                 & 31.0                 & 19.0                 & 33.1                  & 65.8                 & 51.5                 & 36.1                 & 49.2                 \\
TRACE~\cite{guo2024trace}                       & 37.2                 & 21.8                 & 9.6                  & 27.1                  & 43.4                 & 33.9                 & 22.0                 & 32.7                  & 49.7                 & 39.1                 & 28.1                 & 39.0                 \\
TRACE-uni~\cite{guo2024trace}                   & 38.2                 & 22.9                 & 10.4                 & 28.1                  & 44.3                 & 35.1                 & 22.6                 & 33.6                  & 49.9                 & 40.0                 & 29.2                 & 39.8                 \\
TimeSuite~\cite{zeng2024timesuite}              & 56.3                 & 35.5                 & 18.0                 & 38.1                  & 27.1                 & 17.5                 & 8.6                  & 19.8                  & 27.1                 & 16.9                 & 9.9                  & 21.7                 \\
Grounded-VideoLLM~\cite{wang2024grounded_videollm} & 43.3              & 28.7                 & 13.5                 & 30.0                  & 39.2                 & 29.6                 & 19.5                 & 30.0                  & 43.7                 & 33.8                 & 22.5                 & 33.4                 \\
MiMo-VL-7B~\cite{mimo-vl}                       & 57.9                 & 42.6                 & 20.5                 & 39.6                  & 49.3                 & 38.7                 & 22.4                 & 35.5                  & 57.1                 & 42.6                 & 28.4                 & 41.5                 \\
\rowcolor[HTML]{EFEFEF}
Qwen2.5-VL-7B~\cite{qwen2-5-vl}                 & 59.7                 & 37.8                 & 16.6                 & 39.3                  & 44.1                 & 31.0                 & 16.1                 & 31.4                  & 41.5                 & 27.8                 & 15.2                 & 31.6                 \\
\rowcolor[HTML]{E8F0FB}
\textbf{\paper-7B}                  & 70.5                 & 55.6                 & 28.4                 & 48.8                  & 62.8                 & 51.0                 & 32.6                 & 46.2                  & 74.1                 & 62.7                 & 43.1                 & 56.0                 \\

Qwen3-VL-235B-A22B~\cite{Qwen3-VL}             & 71.7                 & 50.8                 & 24.5                 & 47.8                  & 69.0                 & 57.5                 & 39.3                 & 52.2                  & 79.6                 & 70.2                 & 54.5                 & 64.6                 \\
\rowcolor[HTML]{EFEFEF}
Qwen3-VL-8B~\cite{Qwen3-VL}                     & 69.2                 & 53.4                 & 27.5                 & 48.3                  & 62.1                 & 51.2                 & 34.4                 & 46.8                  & 74.2                 & 64.6                 & 49.3                 & 59.4                 \\
\rowcolor[HTML]{E8F0FB}
\textbf{\paper-8B}                  & \textbf{76.6}        & \textbf{63.0}        & \textbf{35.2}        & \textbf{55.2}         & \textbf{68.9}        & \textbf{58.4}        & \textbf{40.6}        & \textbf{53.2}         & \textbf{80.2}        & \textbf{71.6}        & \textbf{55.5}        & \textbf{65.5}        \\ \bottomrule
\end{tabular}%

\caption{
\textbf{Main Results.} We benchmark the performance of various state-of-the-art proprietary and open-source models on \paper-Bench.
Our \paper models are built upon their respective baseline models (preceding rows in the table).
Our \textbf{\paper-7B} not only delivers substantial improvements over the Qwen2.5-VL baseline but also closes the gap with the more powerful Qwen3-VL-8B model.
Building upon the stronger Qwen3-VL baseline, our \textbf{\paper-8B} pushes performance even further, setting a new state-of-the-art among open-source models and surpassing prominent proprietary models like GPT-5 and Gemini-2.5-Flash.
}
\label{tab:main_result}
\end{table*}

\subsection{Empirical Analysis on TimeLens-Bench}
\label{sec:manual_results}


In this section, we present our efforts and findings by applying the above annotation pipeline to existing datasets.
We focus on three most widely-used temporal grounding benchmarks: Charades-STA~\cite{charades-sta}, ActivityNet Captions~\cite{activitynet-captions}, and QVHighlights~\cite{qvhighlights}. These datasets exhibit diversity across video domains, video durations, and query semantics. They are all manually annotated and generally considered the highest-quality VTG datasets available. Therefore, analyzing them offers a representative view of the quality and issues prevalent in existing data.
%
Through diagnosis and refinement, we release \textbf{\paper-Bench}, comprising refined versions of the three aforementioned benchmarks: Charades-TimeLens, ActivityNet-TimeLens, and QVHighlights-TimeLens. Together, they form a comprehensive evaluation suite that combines diversity with high quality.
Detailed statistics for these benchmarks are provided in \cref{sec:bench_details}.

\findingz{1}{Widely-used benchmarks have an alarmingly high proportion of errors.}

\paragraph{Error Statistics and Analysis} As shown in \cref{fig:error_analysis}, we observe an alarmingly high proportion of errors across different categories in these benchmarks. The distribution of error composition varies across different datasets, yet all datasets exhibit consistently high overall error rates. For example, in Charades-STA, we find that 20.6\% of samples violate query uniqueness, while 34.9\% exhibit annotation accuracy issues. Such severe errors will lead to unreliable evaluation results and misguide research efforts.

\paragraph{Qualitative Examples of Errors and Fixes}
As shown in \cref{fig:error_qualitative}, various error examples are identified in existing datasets, including multiple event occurrences, no event occurrence, duplicate queries for the same video, unclear query, and inaccurate annotation. Through our rigorous manual refinement, these detected errors have been properly corrected, significantly improving data quality. Our refined datasets provide more reliable evaluation results.

\findingz{2}{Low-quality evaluation data inflates performance of open-source models, while underestimating proprietary models.}

\paragraph{Counter-intuitive Evaluation Results} 
We evaluate various frontier models on both the original and refined benchmarks, observing \textit{drastically contrasting} performance trends. As illustrated in Fig.~\ref{fig:performance_charades}, on the original benchmarks, we observe a surprising phenomenon: frontier proprietary models like {Gemini-2.5-Pro~\cite{comanici2025gemini-2-5} receive poor scores, whereas open-source models~\cite{wang2025timer1,qwen2-5-vl} attain significantly higher ones.} Conversely, on our refined benchmarks, this trend reverses. The proprietary models exhibit much better results, though with room for improvement, while the open-source models suffer a substantial performance degradation, lagging far behind their proprietary counterparts.
This reversal indicates that the original benchmarks produce \textit{misleading} results due to inherent quality flaws, while our refined benchmarks yield results that align more closely with real-world user experience, providing reliable evaluation for developing better VTG models.

\subsection{Training Data Re-annotation}
By applying our manual pipeline from \cref{sec:manual_pipeline} to a sampled subset of existing VTG training corpus~\cite{didemo,oncescu2021queryd,wang2024cosmo_cap,huang2024vtimellm-internvid-vtime,zala2023hirest}, we found that the training data exhibits an even higher error rate compared to the evaluation benchmarks. This motivated us to refine training data based on scalable re-annotation.
Given the vast scale of the training sets, we employ an automated pipeline to improve their quality based on advanced multimodal models. Owing to the poor quality of these training datasets, especially the high proportion of queries that fail to meet our criteria in \cref{sec:criteria}, we re-annotate the videos rather than refining existing labels. Through this process, we curate \textbf{\paper-100K}, a large-scale, high-quality, and diverse VTG training set. Additional details are provided in \cref{sec:training_data_anno}.

\findingz{3}{Improved annotation quality in training data yields stronger grounding ability.}
As presented in Fig.~\ref{fig:abalation_adding}, models trained on \paper-100K demonstrate substantially improved performance on our refined evaluation benchmarks. This performance gain serves as a direct validation of the data's enhanced quality. Notably, our automated re-annotation for training data is developed entirely independently of the manual benchmark refinement process, ensuring an unbiased evaluation.

\begin{figure*}[t]
  \centering 
  \begin{subfigure}[b]{0.347\textwidth}
    \centering
    \includegraphics[height=4cm,width=\textwidth,keepaspectratio]{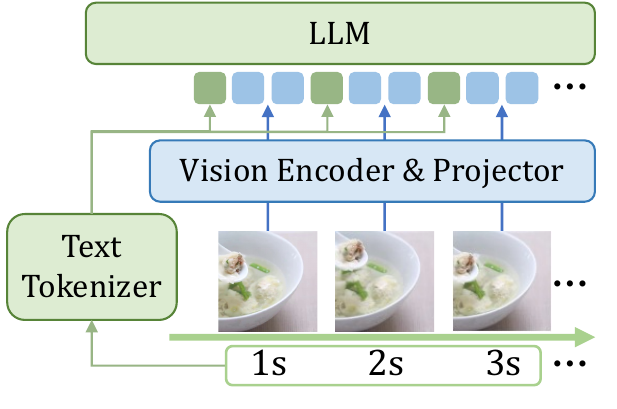}
    \caption{Interleaved Textual Timestamp Encoding.}
    \label{fig:interleaved_textual}
  \end{subfigure}
  \hfill 
  \begin{subfigure}[b]{0.290\textwidth}
    \centering
    \includegraphics[height=4cm,width=\textwidth,keepaspectratio]{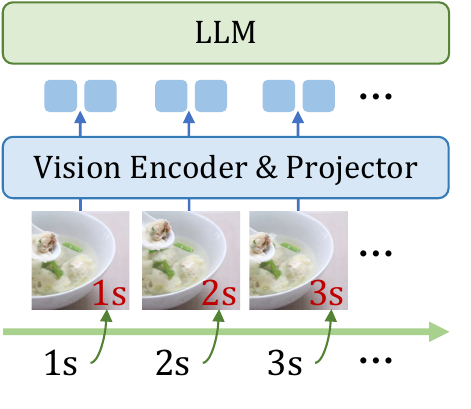}
    \caption{Visual Timestamp Overlay.}
    \label{fig:visual_overlay}
  \end{subfigure}
  \hfill 
  \begin{subfigure}[b]{0.353\textwidth}
    \centering
    \includegraphics[height=4cm,width=\textwidth,keepaspectratio]{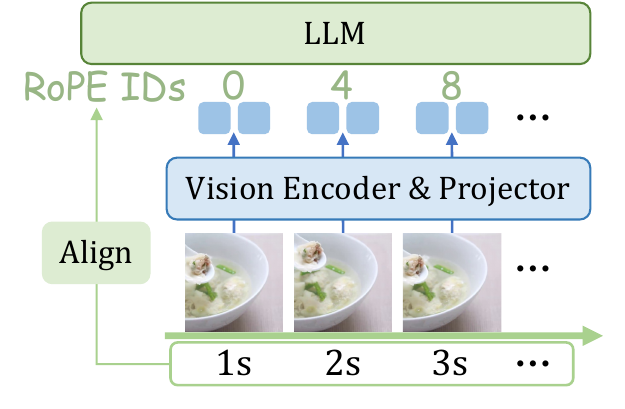}
    \caption{Position-embedding-based Time Encoding.}
    \label{fig:mrope}
  \end{subfigure}

  \caption{\textbf{Illustration of different timestamp encoding schemes.}
    (a) \textbf{Textual Encoding} uses the text tokenizer of LLMs to tokenize textual timestamps into textual tokens.
    (b) \textbf{Visual Overlay} directly overlays timestamps as visual text onto the corresponding frames.
    (c) \textbf{Position-embedding-based Methods} aligns the positional encodings of visual tokens in the LLM with the sampling time of each frame.}
      \label{fig:timestamp_encoding}
\end{figure*}


\begin{table*}[h]
\centering
\small
\setlength{\tabcolsep}{1.15pt}
\renewcommand{\arraystretch}{1.0} 
\begin{tabular}{lc|cccc|cccc|cccc}
\toprule
                                                                                        &                                                                                       & \multicolumn{4}{c}{\textbf{Charades-TimeLens}}                                                                                                                                     & \multicolumn{4}{c}{\textbf{ActivityNet-TimeLens}}                                                                                                                                  & \multicolumn{4}{c}{\textbf{QVHighlights-TimeLens}}                                                                                                            \\ \cmidrule(l){3-14}
\multirow{-2}{*}{\textbf{Method}}                                                       & \multirow{-2}{*}{\textbf{\begin{tabular}[c]{@{}c@{}}Timestamp\\ Format\end{tabular}}} & R1@0.3                                & R1@0.5                                & R1@0.7                                & \multicolumn{1}{c|}{mIoU}                                  & R1@0.3                                & R1@0.5                                & R1@0.7                                & \multicolumn{1}{c|}{mIoU}                                  & R1@0.3                                & R1@0.5                                & R1@0.7                                & mIoU                                  \\ \midrule
Position Embed.~\cite{qwen2-5-vl}                                                       & \multicolumn{1}{c|}{-}                                                                & 57.9                                  & 32.0                                  & 16.9                                  & \multicolumn{1}{c|}{36.6}                                  & 44.8                                  & 31.0                                  & 19.0                                  & \multicolumn{1}{c|}{33.1}                                  & 65.8                                  & 51.5                                  & 36.1                                  & 49.2                                  \\ \midrule
                                                                                        & \multicolumn{1}{c|}{Frame Index}                                                      & 65.5                                  & 48.0                                  & 22.2                                  & \multicolumn{1}{c|}{44.0}                                  & 47.5                                  & 34.0                                  & 17.8                                  & \multicolumn{1}{c|}{33.3}                                  & 61.4                                  & 43.6                                  & 24.1                                  & 42.3                                  \\
\multirow{-2}{*}{Visual Overlay}                                                        & \multicolumn{1}{c|}{Raw Timestamp}                                                    & 67.6                                  & 50.7                                  & 25.7                                  & \multicolumn{1}{c|}{46.3}                                  & 54.0                                  & 42.2                                  & 26.3                                  & \multicolumn{1}{c|}{39.8}                                  & 70.0                                  & 58.3                                  & 42.1                                  & 53.6                                  \\ \midrule
\begin{tabular}[c]{@{}l@{}}Not-Interleaved\\ Textual Prefix\end{tabular}                & \multicolumn{1}{c|}{Raw Timestamp}                                                    & 64.9                                  & 49.4                                  & 27.3                                  & \multicolumn{1}{c|}{45.8}                                  & 48.2                                  & 35.5                                  & 21.4                                  & \multicolumn{1}{c|}{35.2}                                  & 59.7                                  & 45.8                                  & 26.4                                  & 42.8                                  \\ \midrule
                                                                                        & \multicolumn{1}{c|}{Frame Index}                                                      & 66.0                                  & 51.6                                  & 25.6                                  & \multicolumn{1}{c|}{45.6}                                  & 51.0                                  & 39.1                                  & 23.1                                  & \multicolumn{1}{c|}{36.9}                                  & 64.4                                  & 52.1                                  & 32.3                                  & 47.2                                  \\
\multirow{-2}{*}{\begin{tabular}[c]{@{}l@{}}Interleaved \\ Textual Prefix\end{tabular}} & \multicolumn{1}{c|}{\cellcolor[HTML]{C0E4F5}Raw Timestamp}                            & \cellcolor[HTML]{C0E4F5}\textbf{70.0} & \cellcolor[HTML]{C0E4F5}\textbf{53.9} & \cellcolor[HTML]{C0E4F5}\textbf{28.1} & \multicolumn{1}{c|}{\cellcolor[HTML]{C0E4F5}\textbf{48.3}} & \cellcolor[HTML]{C0E4F5}\textbf{57.9} & \cellcolor[HTML]{C0E4F5}\textbf{46.3} & \cellcolor[HTML]{C0E4F5}\textbf{30.5} & \multicolumn{1}{c|}{\cellcolor[HTML]{C0E4F5}\textbf{43.1}} & \cellcolor[HTML]{C0E4F5}\textbf{73.0} & \cellcolor[HTML]{C0E4F5}\textbf{62.2} & \cellcolor[HTML]{C0E4F5}\textbf{46.1} & \cellcolor[HTML]{C0E4F5}\textbf{56.7} \\ \bottomrule
\end{tabular}%
\caption{\textbf{Ablation on timestamp encoding methods.} For each method, we experiment with two timestamp formats: raw timestamps (\eg, ``10.2s'') or frame indices (\eg, ``1, 2, 3''). ``Position Embed.'' means ``Position Embedding''. Results show that \textbf{interleaved textual prefix} with raw timestamps is the most effective approach, while maintaining simplicity.}
\label{tab:time_encode}
\end{table*}

\section{Benchmarking Grounding MLLMs}
In this section, we benchmark the performance of various state-of-the-art proprietary and open-source models on TimeLens-Bench, including our TimeLens models derived from the exploration in \cref{sec:alg_design}.

\paragraph{Evaluation Metrics}
We evaluate VTG performance using the ``R1@m'' metric, which measures the proportion of test instances where the highest-ranked predicted segment achieves an IoU exceeding threshold m (where m takes values from {0.3, 0.5, 0.7}). Additionally, we employ mIoU as a primary measure, computing the mean IoU across the entire test set for conciseness.

\paragraph{Evaluation Results}
As shown in~\cref{tab:main_result}, we observe a significant performance gap between existing open-source and proprietary models, and our \paper models substantially narrow this gap. \paper-7B delivers substantial improvements over its baseline, demonstrating the effectiveness of the insights and best practices obtained from our experiments in~\cref{sec:alg_design}. It surpasses strong open-source competitors such as Time-R1-7B~\cite{wang2025timer1} and MiMo-VL-7B~\cite{mimo-vl}, as well as proprietary models like GPT-4o~\cite{hurst2024gpt-4o} and GPT-5~\cite{OpenAI2025_GPT5}. More remarkably, on the already stronger baseline Qwen3-VL-8B, our \paper-8B model still achieves substantial performance gains, establishing a new state-of-the-art among open-source models and even surpassing frontier proprietary models like Gemini-2.5-Flash~\cite{comanici2025gemini-2-5}.

\section{Exploring Algorithmic Designs}

\label{sec:alg_design}

In this section, we conduct a systematic study on the algorithmic designs for improving MLLMs' VTG performance, covering various aspects from model architectures to training strategies. Leveraging our high-quality training and evaluation suites as a reliable testbed, we derive several novel and valuable insights. As shown in \cref{fig:abalation_adding}, each of our findings contributes a non-trivial performance gain, ultimately culminating in our \textbf{\paper} model.

\vspace{0.5em}
\noindent{\textbf{Experimental Setup.}}
Our experiments use Qwen2.5-VL-7B~\cite{qwen2-5-vl} as the baseline. For RLVR experiments, we employ GRPO~\cite{shao2024deepseekmath} as optimization method. We use \paper-Bench for evaluation and \paper-100K for training.
To ensure rigor, all ablation studies are based on the final, best-performing model configuration,
isolating the impact of a single design choice at a time. Due to limited computational resources, we adopt a lower per-frame resolution  for our ablation experiments. More implementation details are provided in \cref{sec:more_impl_details} of the appendix.

\begin{table*}[t]
\centering
\small
\setlength{\tabcolsep}{1.7pt}
\renewcommand{\arraystretch}{1.15} 
\begin{tabular}{lc|cccc|cccc|cccc}
\toprule
                                                                              &                                                                           & \multicolumn{4}{c|}{\textbf{Charades-TimeLens}}               & \multicolumn{4}{c|}{\textbf{ActivityNet-TimeLens}}            & \multicolumn{4}{c}{\textbf{QVHighlights-TimeLens}}            \\
\multirow{-2}{*}{\begin{tabular}[c]{@{}l@{}}Training\\ Paradigm\end{tabular}} & \multirow{-2}{*}{\begin{tabular}[c]{@{}c@{}}Training\\ Time\end{tabular}} & R1@0.3        & R1@0.5        & R1@0.7        & mIoU          & R1@0.3        & R1@0.5        & R1@0.7        & mIoU          & R1@0.3        & R1@0.5        & R1@0.7        & mIoU          \\ \midrule
SFT (32K Data)                                                                 & 1.0$\times$                                                                     & 68.8          & 53.0          & 26.2          & 47.4          & 53.3          & 42.6          & 27.5          & 39.9          & 65.8          & 54.8          & 40.6          & 52.0          \\ \midrule
SFT (100K Data)                                                                & 2.4$\times$                                                                     & 70.6          & 54.9          & 27.1          & 48.6          & 53.2          & 43.1          & 27.2          & 39.7          & 63.1          & 51.1          & 36.9          & 49.0          \\ \midrule
Thinking-based RLVR                                                           & 1.9$\times$                                                                     & 60.3          & 46.4          & 24.7          & 42.7          & 54.3          & 44.2          & 29.1          & 41.2          & 72.1          & \textbf{62.7}          & \textbf{48.2}          & \textbf{57.8} \\ \midrule
\begin{tabular}[c]{@{}l@{}}SFT + \\ Thinking-free RLVR\end{tabular}           & 2.9$\times$                                                                     & \textbf{71.7} & \textbf{56.7} & \textbf{29.8} & \textbf{50.1} & 56.9           & 46.1 & 30.1 & 42.7 & 72.2           & 60.6           & 43.8          & 55.9          \\ \midrule
\rowcolor[HTML]{C0E4F5}
\textbf{Thinking-free RLVR}                                                    & 1.0$\times$                                                                      & 70.0          & 53.9          & 28.1          & 48.3          & \textbf{57.9} & \textbf{46.3}          & \textbf{30.5} & \textbf{43.1}          & \textbf{73.0} & 62.2 & 46.1 & 56.7          \\ \bottomrule
\end{tabular}%
\caption{\textbf{Ablation on different training paradigms.} We compare the performance and efficiency of different training paradigms, showing that thinking-free RLVR achieves the best performance while maintaining high efficiency. All training is conducted on our quality-improved \paper-100K training data. Training time is measured on 8$\times$ H20 GPUs, where 1.0$\times$ corresponds to approximately 4h10m. As described in~\cref{sec:rlvr_recipes}, before RLVR training, offline inference on the training data is required to select samples with appropriate difficulty; this time is also included in the reported RLVR training time.}
\label{tab:training_paradigm}
\vspace{-0.3em}
\end{table*}

\subsection{Timestamp Encoding}
\label{sec:time_encode}


\findingz{4}{Encoding timestamps as \textit{interleaved textual prefix} is the most effective while maintaining simplicity.}

\noindent To enable MLLMs to perform temporal grounding, a critical design decision is timestamp encoding (\ie, aligning the timestamp of each frame with its corresponding features). Effective timestamp encoding allows the model to accurately perceive the absolute temporal position of each frame and the relative order between frames, thereby producing precise localization results.
As illustrated in~\cref{fig:timestamp_encoding}, various timestamp encoding strategies have been proposed:
\begin{itemize}
    \item \textit{Position-embedding based.} These methods adapt position embeddings in LLMs to represent the temporal position of each frame. For example, MRoPE~\cite{qwen2-5-vl, mimo-vl} and 3D RoPE~\cite{team2025kwai_keye} extend pure-text RoPE to multimodal scenarios, encoding the spatial and temporal dimensions of video frame tokens.

    \item \textit{Visual overlay.} These methods~\cite{ge2025arc-hunyuan-video,wu2025numberit,cheng2025tempura} directly overlay timestamps or frame index onto each frame, enabling MLLMs to ``read'' the temporal position through their OCR capabilities.

    \item \textit{Textual encoding.} These methods convert timestamps into text tokens using the MLLM's text tokenizer. There are two main variants: the \textit{Interleaved} approach~\cite{guo2025seed1_5_vl,li2025unitime,yao2025genS,hong2024cogvlm2,chen2024timemarker} in~\cref{fig:interleaved_textual} inserts timestamp tokens before the visual tokens of each frame. In contrast, the \textit{Non-interleaved} approach~\cite{li2024videochat_flash,wang2025spacevllm,li2023videochat} adds an instruction like ``This video samples $N$ frames of a $T$-second video at $t_1, t_2, \ldots$ seconds.'' into the prompt.
\end{itemize}
%
We conduct a comprehensive comparison of different timestamp encoding methods. For each method, we experiment with two timestamp formats: raw timestamps (\eg, ``10.2s'') or frame indices (\eg, ``1, 2, 3''), which are simpler but neglects the temporal interval between frames. As shown in~\cref{tab:time_encode}, our results reveal: Position-embedding based methods yield unsatisfactory results. Given that they require fundamental modifications to the RoPE mechanism in LLMs, their practicality is limited without large-scale retraining. Instead, interleaved textual prefix with raw timestamps achieves the best performance among all approaches, while remaining simple and intuitive.

\subsection{Optimization Paradigms}

\label{sec:train_paradigm}

\findingz{5}{For the optimization paradigm, a pure thinking-free RLVR approach achieves superior performance and efficiency. Both SFT and thinking-based RLVR are not necessary.}

\noindent In this section, we review different training paradigms and conduct systematic experiments to compare their effectiveness and efficiency for VTG, seeking insights into the optimal training paradigm.

Earlier works~\cite{timechat,huang2024vtimellm-internvid-vtime,zeng2024timesuite,guo2024trace,guo2025vtg-llm} employ supervised fine-tuning (SFT) to improve MLLMs' VTG capability. Recently, some works~\cite{wang2025timer1,tvg-r1} utilize reinforcement learning with verifiable rewards (RLVR), following a ``think-then-answer'' approach~\cite{guo2025deepseek_r1} (details in \cref{sec:more_impl_details}): during sampling, the model first generates an explicit thinking process and then produces the final answer. The task-specific VTG accuracy reward is computed only on the final answer.
Despite these efforts, there lacks a systematic comparison of the respective merits of these methods, leaving some key questions unanswered:
\begin{itemize}
    \item \textbf{Is RLVR superior to SFT?} While the pioneering work Time-R1~\cite{wang2025timer1} demonstrates that RLVR outperforms SFT, they compare the two methods using the same amount of training data, despite RLVR requiring significantly more training time. A fair comparison under equal training budgets remains absent.

    \item \textbf{Is explicit ``thinking'' necessary for RLVR?} Recent works suggest that the thinking process is not essential when applying RLVR to visual perception such as counting~\cite{li2025cls-rl,perception_test}. Whether this holds for VTG, a predominantly perception-oriented task, remains unanswered.

    \item \textbf{Does a preceding SFT phase benefit RLVR?} An SFT phase prior to RLVR is typically employed to enhance the model's capability and facilitate subsequent RLVR training~\cite{team2025kwai_keye, mimo-vl}. However, whether this preceding SFT phase actually improves final performance in the VTG scenario remains unexplored.
\end{itemize}

\noindent In~\cref{tab:training_paradigm}, we compare the performance and efficiency of different training paradigms. Our results reveal that thinking-free RLVR surpasses both SFT and thinking-based RLVR in performance while being more efficient. Adding a preceding SFT phase before RLVR yields no significant performance gain compared to pure RLVR. Overall, a pure thinking-free RLVR approach maintains simplicity, superior performance, and high efficiency.

\begin{figure}[t]
  \centering
   \includegraphics[width=0.98\linewidth]{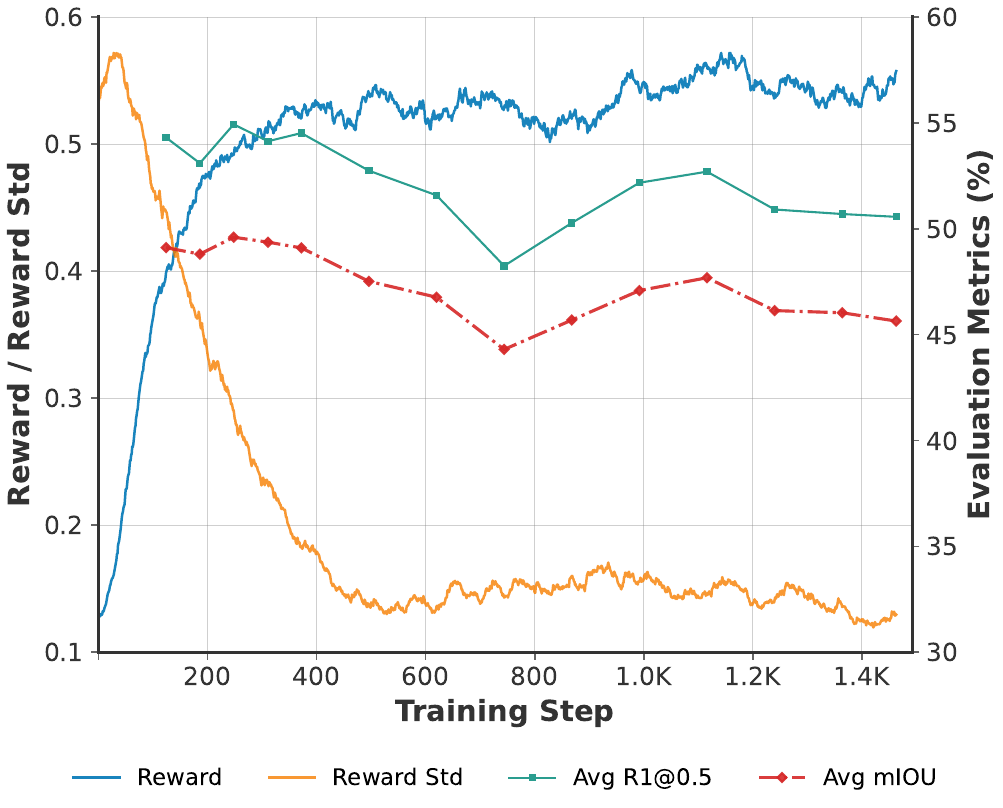}
   \caption{\textbf{The effectiveness of early stopping for RLVR.} We show the trends of training reward and evaluation metrics during RLVR training. When the temporal IoU reward and the within-group reward standard deviation plateau, performance reaches its peak. Continued training beyond this point leads to performance degradation. Therefore, performing \textit{early stopping} when the reward plateaus ensures optimal training efficiency and performance. Training is conducted on $\sim$12K samples selected from \paper-100K via difficulty-aware sampling.}
   \label{fig:early-stop}
\end{figure}

\subsection{Recipes for RLVR Training}
\label{sec:rlvr_recipes}
Building on the finding in~\cref{sec:train_paradigm} that thinking-free RLVR is the optimal training paradigm, in this section, we further explore effective recipes for RLVR training, focusing on two key questions: (i) How long should we train? (ii) How to effectively sample training data?

\findingz{6}{For RLVR training, performing early stopping when reward metrics plateau saves computational cost, while preventing performance degradation.}

\vspace{0.5em}

\noindent{\textbf{How long should we train?}
In SFT, the prevailing wisdom is ``train longer, generalize better''~\cite{hoffer2017train_longer}. Given training data with sufficient scale and quality, we typically train MLLMs for at least one full epoch over the entire dataset, ensuring the model sees as much data as possible to enhance generalization. However, whether this strategy is optimal for RLVR remains to be explored.

In~\cref{fig:early-stop}, we conduct RLVR training on $\sim$12K data from \paper-100K, tracking the reward and evaluating model checkpoints at different training steps on our evaluation benchmarks. When the temporal IoU reward and the within-group reward standard deviation plateau, model performance has reached its peak. Continued training beyond this point leads to performance degradation. Therefore, in RL training, even with sufficiently high data quality, training for a full epoch over all available data is suboptimal. A good practice is performing \textit{early stopping} when reward metrics plateau, which not only saves computational cost but also prevents performance degradation.

\findingz{7}{For RLVR training, sampling training data with sufficiently high difficulty relative to the model is crucial for performance.}
\begin{figure}[t]
  \centering
   \includegraphics[width=0.98\linewidth]{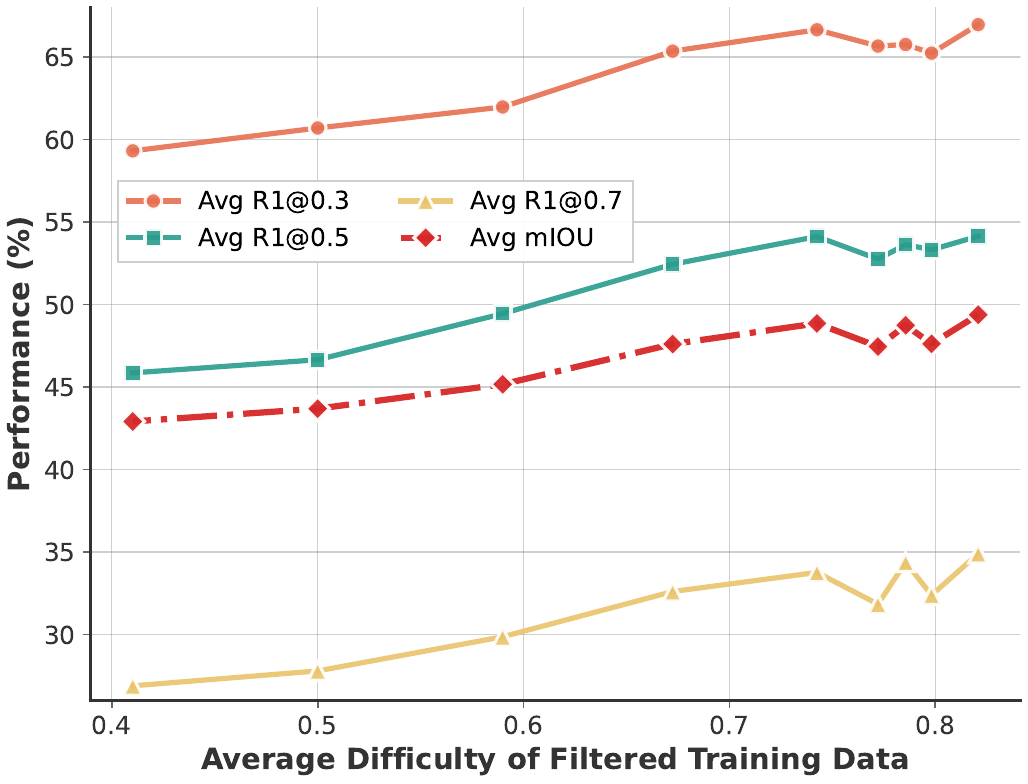}
   \caption{\textbf{The importance of difficulty-based training data sampling for RLVR.} We investigate the impact of training data with different difficulty levels on performance by adjusting the mean of the Gaussian distribution for difficulty-based data sampling. Model performance improves as the average sample difficulty increases, and eventually plateaus when difficulty becomes high. This demonstrates that selecting samples with sufficiently high difficulty is crucial for achieving optimal performance.}
   \label{fig:gaussian_filter}
\end{figure}

\noindent{\textbf{How to sample training data?} For RLVR training, it is crucial to select samples with appropriate difficulty relative to the model, and many works propose assessing training sample difficulty and employing difficulty-aware sampling~\cite{yuan2025vl_cogito,wang2025timer1,hong2025glm_4_5}. To evaluate the impact of sample difficulty on video temporal grounding, we conduct experiments on our \paper-100K high-quality training corpus. Following prior works~\cite{wang2025timer1,yuan2025vl_cogito}, we use the model to be trained to perform offline inference on the training data, compute IoU metrics to estimate sample difficulty, and then perform Gaussian sampling based on sample difficulty (details in \cref{sec:more_impl_details}). By varying the mean of the Gaussian distribution, we obtain training sets with different difficulty levels relative to the model, and conduct RLVR training on each set independently.

As shown in~\cref{fig:gaussian_filter}, model performance improves as the average sample difficulty increases, and eventually plateaus when difficulty becomes sufficiently high (over 0.75). This trend demonstrates that selecting training samples with sufficiently high difficulty relative to the model is crucial for achieving optimal performance.

\clearpage
\noindent \textbf{Acknowledgements.} This work is supported by the National Key R\&D Program of China (No. 2022ZD0160900), the Basic Research Program of Jiangsu (No. BK20250009), the Fundamental and Interdisciplinary Disciplines Breakthrough Plan of the Ministry of Education of China (No. JYB2025XDXM118), and the Collaborative Innovation Center of Novel Software Technology and Industrialization.
{
    \small
    \bibliographystyle{ieeenat_fullname}
    \bibliography{main}
}

\setlength{\abovecaptionskip}{\defaultabovecaptionskip}
\setlength{\belowcaptionskip}{\defaultbelowcaptionskip}
\clearpage
\maketitlesupplementary
\appendix

\section{Conclusion}

In this work, we presented TimeLens, a systematic investigation into building MLLMs with robust video temporal grounding capabilities. On the data front, we first exposed severe quality issues in existing VTG benchmarks. Through meticulous manual re-annotation guided by strict quality criteria, we created TimeLens-Bench, a reliable evaluation suite that dramatically reshapes model rankings and provides trustworthy evaluation for future research. We also developed an automated re-annotation pipeline for noisy training data, yielding TimeLens-100K, a large-scale, high-quality training dataset. On the algorithmic front, our comprehensive exploration yielded several key insights, which culminate in TimeLens models, a family of MLLMs with state-of-the-art VTG performance among open-source models and even surpasses leading proprietary models like GPT-5 and Gemini-2.5-Flash.
By open-sourcing our code, data, and models, we hope TimeLens can serve as a strong foundation for building MLLMs with stronger temporal video grounding capability.

\section{Details of \paper-Bench}
\label{sec:bench_details}

In this section, we provide details on our proposed \paper-Bench, a comprehensive, high-quality evaluation benchmark for video temporal grounding (VTG), comprising refined versions of three mainstream benchmarks: Charades-TimeLens, ActivityNet-TimeLens, and QVHighlights-TimeLens.

\paragraph{Source datasets}
We construct our \paper-Bench based on Charades-STA~\cite{charades-sta}, ActivityNet Captions~\cite{activitynet-captions}, and QVHighlights~\cite{qvhighlights}. For Charades-STA and ActivityNet Captions, we utilize their test splits, while for QVHighlights, we use the validation split, as its test split annotations are not publicly available and prior works~\cite{liu2025videomind,guo2025vtg-llm} also adopt the validation split.
Since the test set of ActivityNet Captions is excessively large, resulting in prohibitively high evaluation cost, we uniformly partition videos based on their duration and sample an equal number of videos from each duration bin. This yields a subset with a video count comparable to Charades-STA and QVHighlights, while maintaining a balanced distribution of video durations.
Although QVHighlights was originally annotated for both video temporal grounding and video highlight detection, our work focuses exclusively on the temporal grounding task.


\begin{table}[h]
\centering
\resizebox{\linewidth}{!}{%
\begin{tabular}{l|cc|ccc|c}
\toprule
\textbf{Dataset}                                 & \# Videos & \begin{tabular}[c]{@{}c@{}}Avg. \\ Duration\end{tabular} & \# Annotations & \begin{tabular}[c]{@{}c@{}}\# Rewritten \\ Queries\end{tabular} & \begin{tabular}[c]{@{}c@{}}\# Refined \\ Time Segments\end{tabular} & Domain     \\ \midrule
Charades-STA~\cite{charades-sta}                 & 1334      & 29.5                                                     & 3720           & -                                                               & -                                                                   & Daily Life \\
\rowcolor[HTML]{E8F0FB} 
Charades-TimeLens                                & 1313      & 29.6                                                     & 3363           & 2467                                                            & 896                                                                 & Daily Life \\ \midrule
ActivityNet Captions~\cite{activitynet-captions} & 4885      & 118.1                                                    & 17031          & -                                                               & -                                                                   & Activity   \\
\rowcolor[HTML]{E8F0FB} 
ActivityNet-TimeLens                             & 1455      & 134.9                                                    & 4500           & 3137                                                            & 1363                                                                & Activity   \\ \midrule
QVHighlights~\cite{qvhighlights}                 & 1529      & 149.6                                                    & 1542           & -                                                               & -                                                                   & Mixed      \\
\rowcolor[HTML]{E8F0FB} 
QVHighlights-TimeLens                            & 1511      & 149.6                                                    & 1541           & 859                                                             & 682                                                                 & Mixed      \\ \midrule
\rowcolor[HTML]{E8F0FB} 
Total (TimeLens-Bench)                           & 4279      & 107.8                                                    & 9404           & 6463                                                            & 2941                                                                & Mixed      \\ \bottomrule
\end{tabular}%
}
\caption{Statistics of the datasets in our proposed TimeLens-Bench, compared against their original versions (Charades-STA~\cite{charades-sta}, ActivityNet Captions~\cite{activitynet-captions} and QVHighlights~\cite{qvhighlights}).
}
\label{tab:bench-statistics}
\end{table}

\paragraph{Detailed Statistics}
In \cref{tab:bench-statistics}, we present detailed statistics of our \paper-Bench and its source dataset counterparts. For each source dataset, annotations with high-quality queries had their corresponding temporal segments refined. For the remaining annotations with low-quality queries, we either revised the queries or rewrote them entirely. A small fraction of queries that were deemed unfixable were subsequently discarded. Overall, \paper-Bench comprises a total of 4,279 videos with an average duration of 107.8 seconds, and 9,404 annotations.

\subsection{Evaluation Metrics}
We evaluate model performance on \paper-Bench using four standard metrics: Recall@1 at IoU thresholds of 0.3, 0.5, and 0.7 (denoted as R1@0.3, R1@0.5, R1@0.7), and mean Intersection over Union (mIoU).
\begin{itemize}
    \item \textbf{mIoU} is defined as the average of the temporal Intersection over Union (IoU) scores between the predicted and ground-truth segments across all test samples.
    \item \textbf{R1@$m$} measures the percentage of samples for which the temporal IoU of the prediction exceeds a given threshold $m$.
\end{itemize}

While \paper-Bench can be treated as a single unified benchmark for computing the aforementioned metrics, we compute and report metrics \textbf{separately} on its three constituent benchmarks to enable a more fine-grained analysis of model performance across different domains. We encourage future works to adopt this evaluation protocol.

\begin{table*}[t]
\centering
\resizebox{\textwidth}{!}{%
\begin{tabular}{lcccc|cccc|cccc}
\toprule
                                         & \multicolumn{4}{c|}{\textbf{Charades-TimeLens}}               & \multicolumn{4}{c|}{\textbf{ActivityNet-TimeLens}}            & \multicolumn{4}{c}{\textbf{QVHighlights-TimeLens}}            \\
\multirow{-2}{*}{\textbf{Training Data}} & R1@0.3        & R1@0.5        & R1@0.7        & mIoU          & R1@0.3        & R1@0.5        & R1@0.7        & mIoU          & R1@0.3        & R1@0.5        & R1@0.7        & mIoU          \\ \midrule
Original Noisy Data                      & 52.6          & 30.4          & 14.0          & 35.6          & 45.0          & 29.5          & 16.0          & 31.3          & 61.3          & 46.1          & 29.1          & 44.6          \\
\rowcolor[HTML]{E8F0FB} 
\textbf{\paper-100K}                     & \textbf{70.0} & \textbf{53.9} & \textbf{28.1} & \textbf{48.3} & \textbf{57.9} & \textbf{46.3} & \textbf{30.5} & \textbf{43.1} & \textbf{73.0} & \textbf{62.2} & \textbf{46.1} & \textbf{56.7} \\ \bottomrule
\end{tabular}%
}
\caption{\textbf{Ablation on training data.} Our \paper-100K training set significantly improves model performance, validating its enhanced quality.}
\label{tab:train_data}
\end{table*}

\begin{table*}[h]
\centering
\resizebox{\linewidth}{!}{%
\begin{tabular}{lcccc|cccc|cccc}
\toprule
                                           & \multicolumn{4}{c|}{\textbf{Charades-TimeLens}}               & \multicolumn{4}{c|}{\textbf{ActivityNet-TimeLens}}            & \multicolumn{4}{c}{\textbf{QVHighlights-TimeLens}}            \\
\multirow{-2}{*}{\textbf{Model}}           & R1@0.3        & R1@0.5        & R1@0.7        & mIoU          & R1@0.3        & R1@0.5        & R1@0.7        & mIoU          & R1@0.3        & R1@0.5        & R1@0.7        & mIoU          \\ \midrule
Qwen2.5-VL-3B~\cite{qwen2-5-vl} (Baseline) & 51.3          & 30.5          & 14.2          & 33.9          & 25.3          & 18.2          & 9.4           & 18.4          & 27.4          & 19.3          & 10.6          & 20.9          \\
\rowcolor[HTML]{E8F0FB} 
\quad+\textbf{\paper}                         & {\ul 63.5}    & {\ul 48.1}    & {\ul 23.9}    & {\ul 43.3}    & {\ul 56.6}    & {\ul 44.9}    & {\ul 27.4}    & {\ul 41.2}    & {\ul 71.9}    & {\ul 58.9}    & {\ul 37.8}    & {\ul 52.9}    \\
Qwen2.5-VL-7B~\cite{qwen2-5-vl} (Baseline) & 59.7          & 37.8          & 16.6          & 39.3          & 44.1          & 31.0          & 16.1          & 31.4          & 41.5          & 27.8          & 15.2          & 31.6          \\
\rowcolor[HTML]{E8F0FB} 
\quad+\textbf{\paper}                         & \textbf{70.5} & \textbf{55.6} & \textbf{28.4} & \textbf{48.8} & \textbf{62.8} & \textbf{51.0} & \textbf{32.6} & \textbf{46.2} & \textbf{74.1} & \textbf{62.7} & \textbf{43.1} & \textbf{56.0} \\ \bottomrule
\end{tabular}%
}
\caption{\textbf{Results across different model sizes.} The best and second-best results are highlighted in \textbf{bold} and \underline{underlined}, respectively.}
\label{tab:3b_results}
\end{table*}
\begin{table*}[t]
\centering
\resizebox{0.8\linewidth}{!}{%
\begin{tabular}{lcccc|cccc}
\toprule
                                                & \multicolumn{4}{c|}{\textbf{Charades-TimeLens}}               & \multicolumn{4}{c}{\textbf{Charades-STA}} \\
\multirow{-2}{*}{\textbf{Model}}                & R1@0.3        & R1@0.5        & R1@0.7        & mIoU          & R1@0.3    & R1@0.5   & R1@0.7   & mIoU    \\ \midrule
\textit{Proprietary Models}                     &               &               &               &               &           &          &          &         \\
GPT-4o~\cite{hurst2024gpt-4o}                   & 60.6          & 44.5          & 23.5          & 41.8          & 51.6      & 27.9     & 11.7     & 34.7    \\
GPT-5~\cite{OpenAI2025_GPT5}                    & 59.3          & 42.0          & 22.0          & 40.5          & 39.7      & 18.3     & 6.2      & 28.4    \\
Gemini-2.0-Flash~\cite{comanici2025gemini-2-5}  & 66.4          & 53.5          & 27.1          & 46.7          & 55.6      & 29.0     & 9.5      & 35.1    \\
Gemini-2.5-Flash~\cite{comanici2025gemini-2-5}  & 68.7          & 56.1          & 30.6          & 48.6          & 47.0      & 21.8     & 7.1      & 31.1    \\
Gemini-2.5-Pro~\cite{comanici2025gemini-2-5}    & 74.1          & 61.1          & 34.0          & 52.8          & 53.9      & 25.5     & 8.8      & 34.6    \\ \midrule
\textit{Open-Source Models}                     &               &               &               &               &           &          &          &         \\
VideoChat-Flash-7B~\cite{li2024videochat_flash} & 60.2          & 37.9          & 17.8          & 39.7          & 72.5*     & 51.4*    & 26.4*    & 45.2*   \\
VideoChat-R1-7B~\cite{li2025videochat_r1}       & 51.9          & 30.8          & 11.7          & 33.7          & -         & 71.7*    & 50.2*    & 60.8*   \\
Time-R1-7B~\cite{wang2025timer1}                & 57.9          & 32.0          & 16.9          & 36.6          & 78.1*     & 60.8*    & 35.3*    & 58.1*   \\
MiMo-VL-7B~\cite{mimo-vl}                       & 57.9          & 42.6          & 20.5          & 39.6          & -         & -        & -        & 50.0*   \\
Qwen2.5-VL-7B~\cite{qwen2-5-vl} (Baseline)      & 59.7          & 37.8          & 16.6          & 39.3          & 59.4      & 38.2     & 18.1     & 43.6*   \\
\rowcolor[HTML]{E8F0FB} 
\textbf{\paper-7B}                              & \textbf{70.5} & \textbf{55.6} & \textbf{28.4} & \textbf{48.8} & 70.7      & 39.8     & 14.5     & 42.3    \\ \bottomrule
\end{tabular}%
}
\caption{\textbf{Results on our refined Charades-\paper and the original Charades-STA benchmark.}    * indicates results from the original paper, other results are from our evaluation.}
\label{tab:old_new_charades}
\end{table*}

\section{More Implementation Details.}
\label{sec:more_impl_details}

\begin{table*}[t]
\centering
\resizebox{0.8\textwidth}{!}{%
\begin{tabular}{lcccc|cccc}
\toprule
                                               & \multicolumn{4}{c|}{\textbf{ActivityNet-TimeLens}}                                                                                                            & \multicolumn{4}{c}{\textbf{ActivityNet-Captions}} \\
\multirow{-2}{*}{\textbf{Model}}               & R1@0.3                                & R1@0.5                                & R1@0.7                                & mIoU                                  & R1@0.3      & R1@0.5     & R1@0.7     & mIoU      \\ \midrule
\textit{Proprietary Models}                    &                                       &                                       &                                       &                                       &             &            &            &           \\
Gemini-2.0-Flash~\cite{comanici2025gemini-2-5} & 62.9                                  & 54.0                                  & 37.7                                  & 49.3                                  & 50.4        & 33.2       & 19.9       & 36.5      \\
Gemini-2.5-Flash~\cite{comanici2025gemini-2-5} & 66.8                                  & 57.5                                  & 41.3                                  & 52.5                                  & 51.2        & 34.7       & 21.0       & 37.4      \\ \midrule
\textit{Open-Source Models}                    &                                       &                                       &                                       &                                       &             &            &            &           \\
VideoChat-R1-7B~\cite{li2025videochat_r1}      & 35.0                                  & 23.9                                  & 11.3                                  & 25.0                                  & -           & 33.3*      & 16.7*      & 35.5*     \\
Time-R1-7B~\cite{wang2025timer1}               & 44.8                                  & 31.0                                  & 19.0                                  & 33.1                                  & 58.1*       & 39.0*      & 21.4*      & 40.5*     \\
MiMo-VL-7B~\cite{mimo-vl}                      & 49.3                                  & 38.7                                  & 22.4                                  & 35.5                                  & 39.3        & 24.3       & 12.9       & 28.1      \\
Qwen2.5-VL-7B~\cite{qwen2-5-vl} (Baseline)     & 44.1                                  & 31.0                                  & 16.1                                  & 31.4                                  & 34.5        & 20.8       & 11.2       & 26.4      \\
\rowcolor[HTML]{E8F0FB} 
\textbf{\paper-7B}                             & \cellcolor[HTML]{E8F0FB}\textbf{62.8} & \cellcolor[HTML]{E8F0FB}\textbf{51.0} & \cellcolor[HTML]{E8F0FB}\textbf{32.6} & \cellcolor[HTML]{E8F0FB}\textbf{46.2} & 53.5        & 35.2       & 19.7       & 37.7      \\ \bottomrule
\end{tabular}%
}
\caption{\textbf{Results on our refined ActivityNet-\paper and the original ActivityNet-Captions benchmark.}    * indicates results from the original paper, other results are from our evaluation.}
\label{tab:old_new_anet}
\end{table*}
\begin{table*}[h]
\centering
\resizebox{0.75\linewidth}{!}{%
\begin{tabular}{lcccc|llll}
\toprule
                                               & \multicolumn{4}{c|}{\textbf{\begin{tabular}[c]{@{}c@{}}QVHighlights-TimeLens\end{tabular}}}                                                                & \multicolumn{4}{c}{\textbf{QVHighlights}}                                                                       \\
\multirow{-2}{*}{\textbf{Model}}               & R1@0.3                                & R1@0.5                                & R1@0.7                                & mIoU                                  & \multicolumn{1}{c}{R1@0.3} & \multicolumn{1}{c}{R1@0.5} & \multicolumn{1}{c}{R1@0.7} & \multicolumn{1}{c}{mIoU} \\ \midrule
\textit{Proprietary}                           &                                       &                                       &                                       &                                       & \multicolumn{1}{c}{}       & \multicolumn{1}{c}{}       & \multicolumn{1}{c}{}       & \multicolumn{1}{c}{}     \\
Gemini-2.0-Flash~\cite{comanici2025gemini-2-5} & 76.2                                  & 66.4                                  & 48.3                                  & 60.8                                  & 72.1                       & 58.1                       & 41.9                       & 54.9                     \\
Gemini-2.5-Flash~\cite{comanici2025gemini-2-5} & 78.2                                  & 69.4                                  & 55.0                                  & 64.3                                  & 76.9                       & 62.6                       & 46.7                       & 59.1                     \\ \midrule
\textit{Open-Source}                           &                                       &                                       &                                       &                                       & \multicolumn{1}{c}{}       & \multicolumn{1}{c}{}       & \multicolumn{1}{c}{}       & \multicolumn{1}{c}{}     \\
MiMo-VL-7B~\cite{mimo-vl}                      & 57.1                                  & 42.6                                  & 28.4                                  & 41.5                                  & 59.6                       & 43.3                       & 24.7                       & 41.8                     \\
Qwen2.5-VL-7B~\cite{qwen2-5-vl}                & 41.5                                  & 27.8                                  & 15.2                                  & 31.6                                  & 29.2                       & 19.2                       & 11.9                       & 27.4                     \\
\rowcolor[HTML]{E8F0FB} 
\textbf{\paper-7B}                             & \cellcolor[HTML]{E8F0FB}\textbf{74.1} & \cellcolor[HTML]{E8F0FB}\textbf{62.7} & \cellcolor[HTML]{E8F0FB}\textbf{43.1} & \cellcolor[HTML]{E8F0FB}\textbf{56.0} & 78.4                       & 63.6                       & 43.7                       & 57.4                     \\ \bottomrule
\end{tabular}%
}
\caption{\textbf{Results on our refined QVHighlights-\paper and the original QVHighlights benchmark.}  All results are from our evaluation.}
\label{tab:old_new_qvhighlights}
\end{table*}
\begin{figure*}[t]
  \centering
  \begin{subfigure}[b]{0.9\linewidth}
     \centering
     \includegraphics[width=\textwidth]{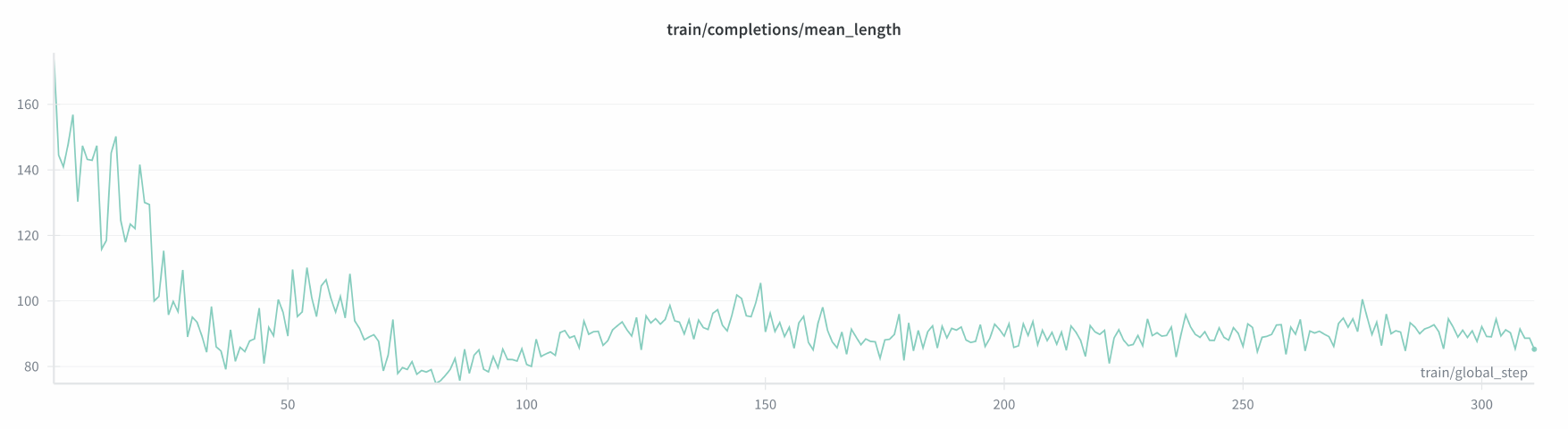}
     \caption{}
  \end{subfigure}
  \vspace{1em} 
  \begin{subfigure}[b]{0.9\linewidth}
     \centering
     \includegraphics[width=\textwidth]{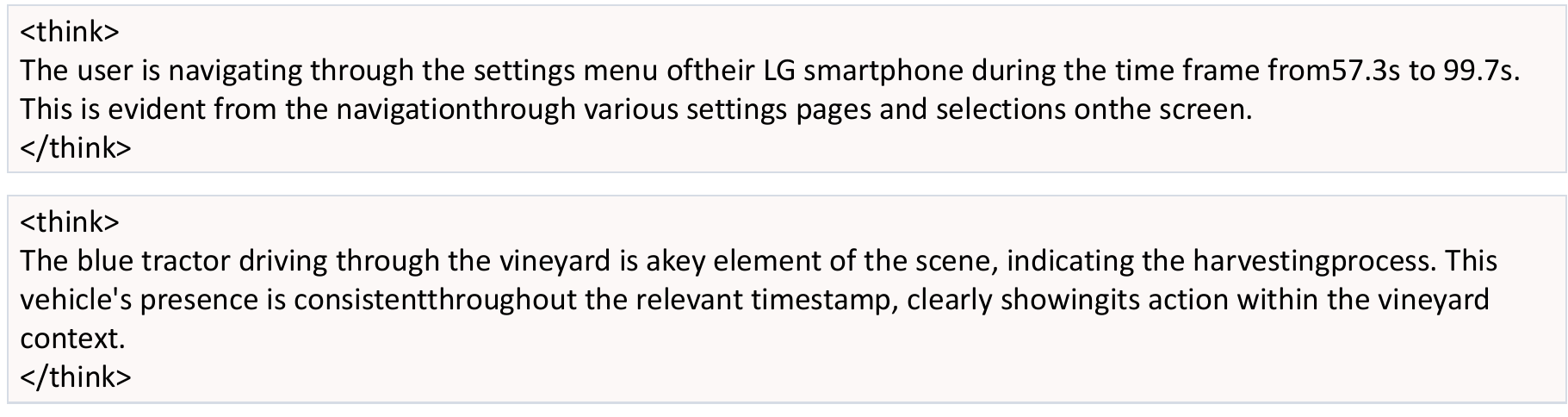}
     \caption{}
  \end{subfigure}

  \caption{\textbf{Observations during thinking-based RLVR training.}
  (a) The model's thinking length gradually decreases during training.
  (b) The model generates simple thinking content, primarily perception-related, without exhibiting any complex reasoning.}
  \label{fig:thinking}
\end{figure*}

\subsection{Preliminaries for RL}

\paragraph{Thinking-based vs. Thinking-free RLVR.}
We formalize the distinction between thinking-based and thinking-free RLVR paradigms using GRPO~\cite{shao2024deepseekmath} as the reinforcement learning algorithm.

In the task of video temporal grounding, given a video~$v$ and query~$q$, the model generates a response~$y$. For GRPO training, for each input pair~$(v, q)$ in the training set~$D$, we sample a group of~$G$ responses~$\{y^{(i)}\}_{i=1}^G$ from the policy~$\pi_\theta$, compute their rewards~$\{r(y^{(i)})\}_{i=1}^G$, and optimize the policy to maximize the relative advantage within the group:
\begin{equation}
    \mathcal{L}_{\text{GRPO}} = -\mathbb{E}_{(v,q) \sim \mathcal{D}} \mathbb{E}_{y^{(i)} \sim \pi_\theta} \left[ A^{(i)} \log \pi_\theta(y^{(i)} | v, q) \right],
\end{equation}
where the advantage is computed as:
\begin{equation}
    A^{(i)} = r(y^{(i)}) - \frac{1}{G}\sum_{j=1}^G r(y^{(j)}).
\end{equation}

The key distinction between the two paradigms lies in the response structure and reward computation. In \textbf{thinking-based RLVR}, following the ``think-then-answer'' approach~\cite{guo2025deepseek_r1}, the response consists of two parts:
\begin{equation}
    y = [y_{\text{thinking}}, y_{\text{answer}}],
\end{equation}
where $y_{\text{thinking}}$ represents the explicit reasoning process and $y_{\text{answer}}$ contains the predicted temporal segment~$\hat{S} = (\hat{t}_{\text{start}}, \hat{t}_{\text{end}})$. The reward function combines accuracy and format compliance:
\begin{equation}
    r(y) = r_{\text{acc}}(y_{\text{answer}}) + r_{\text{format}}(y),
\end{equation}
where $r_{\text{acc}}(y_{\text{answer}}) = \text{IoU}(\hat{S}, S^*)$ measures the temporal intersection-over-union with the ground truth segment~$S^*$, and $r_{\text{format}}(y)$ ensures proper output formatting following the ``think-then-answer'' structure.

In contrast, our \textbf{thinking-free RLVR} directly generates the answer without explicit reasoning:
\begin{equation}
    y = y_{\text{answer}},
\end{equation}
with a simplified reward based solely on grounding accuracy:
\begin{equation}
    r(y) = r_{\text{acc}}(y) = \text{IoU}(\hat{S}, S^*).
\end{equation}

As shown in~\cref{tab:training_paradigm}, the thinking-free paradigm eliminates the need for explicit reasoning generation and format reward engineering, leading to simpler mplementation, faster training and inference, and superior performance.

\paragraph{Difficulty-aware Sampling}
To formalize difficulty-aware sampling~\cite{yuan2025vl_cogito,wang2025timer1,hong2025glm_4_5}, we first perform offline inference with the model to be trained on the training dataset~$\mathcal{D} = \{(v_i, q_i, S_i^*)\}_{i=1}^N$. For each sample, we obtain the predicted segment~$\hat{S}_i$ and compute the difficulty estimate as:
\begin{equation}
    d_i = 1 - \text{IoU}(\hat{S}_i, S_i^*),
\end{equation}
where higher values indicate more challenging samples for the current model.

We then compute sampling weights for each sample based on its difficulty. Following ~\cite{yuan2025vl_cogito,wang2025timer1}, we employ Gaussian sampling to construct a training subset where samples with difficulty around a target mean~$\mu$ are more likely to be selected. Let~$g(d; \mu, \sigma^2)$ denote the target Gaussian distribution:
\begin{equation}
    g(d; \mu, \sigma^2) = \frac{1}{\sqrt{2\pi\sigma^2}} \exp\left(-\frac{(d-\mu)^2}{2\sigma^2}\right).
\end{equation}

To ensure that samples with difficulty~$d$ are selected with probability proportional to~$g(d)$, we compute the sampling weight for each sample~$i$ as:
\begin{equation}
    w_i = \frac{g(d_i; \mu, \sigma^2)}{\hat{p}(d_i)},
\end{equation}
where $\hat{p}(d_i)$ is the empirical density of samples with difficulty~$d_i$ in the original dataset. This density correction ensures that the difficulty distribution of the sampled subset follows the target Gaussian distribution, rather than being biased by the original difficulty distribution in~$\mathcal{D}$.

By varying the mean~$\mu$ of the Gaussian distribution, we obtain training sets with different average difficulty levels and conduct RLVR training on each independently to evaluate the impact of sample difficulty on final model performance in~\cref{sec:rlvr_recipes}.

\subsection{Experimental Setup}
\label{sec:exp_setup}

Unless otherwise specified, all experiments are conducted using Qwen2.5-VL-7B~\cite{qwen2-5-vl} as the base model. Under the Qwen2.5-VL architecture, every two consecutive video frames are merged during the patch-embedding stage of the vision encoder.

\paragraph{Model Configuration}
We sample video frames at 2 FPS. For all ablation experiments, we set the minimum number of tokens per frame (i.e., every two merged frames) to $\texttt{min\_tokens}=16$, and the maximum number of tokens for the entire video to $\texttt{total\_tokens}=3584$. Under this configuration, the model adaptively adjusts the spatial resolution based on the video's duration and raw resolution. For the final \paper models presented in ~\cref{tab:main_result}, we scale the resolution budget to $\texttt{min\_tokens}=64$ and $\texttt{total\_tokens}=14336$. Under this setting, for a 110-second video, the maximum resolution per frame is about $320 \times 320$ pixels.

\paragraph{Timestamp Encoding}
In~\cref{tab:time_encode} and~\cref{sec:time_encode}, we experiment with different timestamp encoding methods. For \textit{position-embedding} methods, we directly adopt the original MRoPE implementation from Qwen2.5-VL~\cite{qwen2-5-vl}. For \textit{Visual Overlay}, we render timestamps in red with a font size of $40$\,pt, overlaid at the bottom-left corner of each frame. For \textit{Non-interleaved Textual Encoding}, an instruction like ``This video samples $N$ frames of a $T$-second video at $t_1, t_2, \ldots$ seconds'' is prepended to the prompt.

For \textit{Interleaved Textual Encoding}, timestamps are converted to text with one decimal place retained (\eg, 10.2 seconds), then tokenized and prepended to the tokens of each frame. As described above, when processing videos, every two consecutive frames are merged during the patch-embedding stage of the vision encoder, while for images, each image is duplicated into two identical copies for merging. Since we insert textual timestamp tokens into the original video tokens to form an interleaved visual-text sequence, we treat each video frame as an independent image and duplicate it for processing. This approach allows us to bypass the original MRoPE mechanism entirely, enabling an \textit{isolated} comparison between Interleaved Textual Encoding and MRoPE. Meanwhile, we adopt 1 FPS sampling to ensure the computational cost matches that of other temporal encoding methods using 2 FPS sampling.

\paragraph{Training Configuration}
For all training procedures, we freeze the vision encoder while updating all other parameters, and train for one epoch. For supervised fine-tuning (SFT) experiments, we use a batch size of $128$ and a learning rate of $1 \times 10^{-5}$. For reinforcement learning (RL) experiments, we perform difficulty-aware data sampling with a Gaussian distribution where $\mu = 0.05$ and $\sigma = 0.2$. The training batch size is $8$, each prompt samples $8$ roll-outs, the learning rate is $1 \times 10^{-6}$, and the KL coefficient $\beta$ is set to $0$. We train until we observe that the reward plateaus and then perform early stopping. In practice, this corresponds to approximately 310 training steps ($\sim$2.5K training examples) for Qwen2.5-VL models.

Throughout the development of this work, our experiments were conducted based on Qwen2.5-VL. More recently, the more powerful Qwen3-VL~\cite{Qwen3-VL} models have been released, so we also validated the effectiveness of our data and recipe on Qwen3-VL. We observed that directly applying RL training on Qwen3-VL fails to yield improvements, likely because Qwen3-VL has undergone large-scale multi-task RL training that includes VTG data, preventing the model from generating rollouts with sufficient diversity on VTG task during our continual RL. Therefore, we first perform a small SFT stage to, in a sense, revert the model back to the ``base model'' state before RL. This is merely a workaround specific to Qwen3-VL, a model that has already acquired strong VTG capabilities through an RL stage similar to that proposed in this paper. In the common scenario, our recipes are designed to enhance the VTG capabilities of a ``base MLLM'', where this trick is not required.

\section{More Experimental Results}

\paragraph{Results across different model sizes}
In~\cref{tab:3b_results}, we demonstrate the effectiveness of our proposed design principles across various model sizes.
Across base models of varying sizes, \paper consistently delivers significant performance gains.
Remarkably, despite having fewer parameters, \paper-3B substantially surpasses even the larger Qwen-2.5-VL-7B model.

\paragraph{Generalization to an external benchmark}
To further validate that our recipe generalizes beyond \paper-Bench, we additionally evaluate on VUE-TR, a high-quality human-annotated VTG benchmark introduced concurrently in Vidi~\cite{team2025vidi}. As shown in~\cref{tab:vue_tr}, \paper-7B achieves the best result among the compared models, indicating that the gains from our data curation and algorithmic recipe are not limited to our own refined benchmark suite.
\begin{table}[h]
\centering
\small
\setlength{\tabcolsep}{6pt}
\begin{tabular}{lc}
\toprule
\textbf{Model} & \textbf{IoU @ VUE-TR (0-200s)} \\ \midrule
Qwen2.5-VL-7B~\cite{qwen2-5-vl} & 36.0 \\
GPT-4o~\cite{hurst2024gpt-4o} & 34.5 \\
Gemini-2.5-Pro~\cite{comanici2025gemini-2-5} & 41.6 \\
\rowcolor[HTML]{E8F0FB}
\textbf{\paper-7B} & \textbf{45.1} \\ \bottomrule
\end{tabular}
\caption{\textbf{Results on the VUE-TR benchmark from Vidi~\cite{team2025vidi}.} \paper-7B achieves the best IoU among the compared models on this external benchmark.}
\label{tab:vue_tr}
\end{table}

\paragraph{Comparison of results on \paper-Bench and original benchmarks}
In~\cref{tab:old_new_charades},~\cref{tab:old_new_anet}, and~\cref{tab:old_new_qvhighlights}, we compare the evaluation results of various models on \paper-Bench and the original benchmarks. On the original benchmarks, due to data quality issues, open-source models \textit{deceptively} surpass state-of-the-art proprietary models like Gemini-2.5-Pro. On our refined benchmarks, model capabilities are more reliably evaluated, with proprietary models maintaining a significant advantage over open-source models. Remarkably, our \paper model substantially narrows the performance gap between open-source and proprietary models.

\paragraph{Results on general video understanding}
In~\cref{tab:video_qa}, we evaluate \paper-7B's general video understanding capabilities on VideoMME~\cite{fu2025videomme}, the most comprehensive and widely-adopted video understanding benchmark.
The results demonstrate that \paper-7B maintains the strong general video understanding capability of its base model.
This validates that our proposed design principles can effectively enhance video temporal grounding capabilities without sacrificing general-purpose video understanding abilities.
\begin{table}[h]
\centering
\resizebox{\linewidth}{!}{%
\begin{tabular}{l|cccc}
\toprule
                                           & \multicolumn{4}{c}{Video-MME}                                                                                                                                            \\ \cmidrule(l){2-5} 
\multirow{-2}{*}{Model}                    & Short                                      & Medium                                     & Long                              & All                                        \\ \midrule
\rowcolor[HTML]{EFEFEF} 
Qwen2.5-VL-7B~\cite{qwen2-5-vl} (Baseline) & 64.3\textsuperscript{*}                    & 75.2\textsuperscript{*}                    & \textbf{55.1}\textsuperscript{*}  & 65.1\textsuperscript{\dag}                 \\
\rowcolor[HTML]{E8F0FB} 
\textbf{\paper-7B}                        & \textbf{66.4\phantom{\textsuperscript{*}}} & \textbf{76.7\phantom{\textsuperscript{*}}} & 54.1\phantom{\textsuperscript{*}} & \textbf{65.7\phantom{\textsuperscript{*}}} \\ \bottomrule
\end{tabular}%
}
\caption{\textbf{Results on the general video understanding benchmark Video-MME~\cite{fu2025videomme}.} The results show that \paper-Qwen2.5-VL-7B maintains strong general video understanding capability while achieving substantial improvements in video temporal grounding. \textsuperscript{*} Results reproduced by us. \textsuperscript{\dag} Results reported in original paper.}
\label{tab:video_qa}
\end{table}

\section{Discussion on Thinking-free vs. Thinking-based RLVR}
We analyze the possible reasons why thinking-based RLVR underperforms thinking-free methods in our experiments, from both intuitive and empirical perspectives.

\paragraph{Intuitive Analysis.}
When manually examining and refining existing grounding datasets, we observe that queries in the grounding task are relatively straightforward, primarily testing the model's perception capability: whether the model can accurately localize the corresponding event in a long video containing massive information. From a human perspective, completing existing grounding tasks indeed relies mainly on intuition and instinct, rather than complex reasoning.

\paragraph{Empirical Observations.}
In our experiments, when training with thinking-based RLVR, the model's thinking length gradually decreases and converges to simple, non-reasoning thinking processes, as shown in~\cref{fig:thinking}. This suggests that the model learns to bypass explicit reasoning when it provides no benefit to the task.

\paragraph{Implications and Future Work.}
We believe that most samples in existing video temporal grounding data do not require complex reasoning capabilities, but rather demand sufficiently robust long-video perception and localization abilities. Due to the high cost and corresponding low quality of existing data annotation, as well as limitations in current algorithmic designs, existing MLLMs cannot yet achieve this perfectly. Therefore, in this work, we focus on addressing these two fundamental issues. Meanwhile, we believe that certain grounding tasks do require reasoning capabilities, and we leave the exploration of reasoning-intensive VTG scenarios to future work.

\section{Additional Discussions}

\paragraph{Why do VideoChat-R1 and Time-R1 behave differently across benchmarks?}
VideoChat-R1~\cite{li2025videochat_r1} is trained on single-domain data with relatively limited quality control, and it does not incorporate the training recipes that we find most effective for VTG, such as difficulty-aware sampling and early stopping. As a result, it underperforms even the Qwen2.5-VL baseline on our refined benchmarks. Time-R1~\cite{wang2025timer1} by contrast, benefits from a stronger RL-oriented training recipe and therefore transfers better to ActivityNet-TimeLens and QVHighlights-TimeLens, where long-video localization is a major challenge. However, its training data is still noisier than ours, which likely limits temporal precision and makes it less competitive on Charades-TimeLens, where videos are shorter but localization accuracy must be much higher.

\paragraph{Training setting for timestamp encoding.}
All methods in~\cref{tab:time_encode} are compared after RLVR training rather than in a training-free setting. This controlled setup is important because the temporal encoding method interacts with the optimization paradigm: the final comparison reflects how well each representation supports post-training for VTG, instead of only measuring zero-shot prompting behavior.

\paragraph{Early stopping.}
In our single-task VTG setting, reward plateauing is a reliable signal that continued RLVR is unlikely to improve grounding quality and may even degrade it, so early stopping saves computation while preventing over-training. We regard this as a practical recipe rather than a universal rule. In more general multi-task training, a single stopping criterion may be harder to define because different capabilities can peak at different stages, requiring more sophisticated recipes (e.g., remove training data for a single task when reward on this task peaks, while continuing training for other tasks).

\section{Annotation Interface and Manual}
\label{seg:anno_tool}

We present our annotation interface in~\cref{fig:anno_interface} and our annotation manual in~\cref{fig:anno_manual}.

\begin{figure*}[t]
  \centering
   \includegraphics[width=1\linewidth]{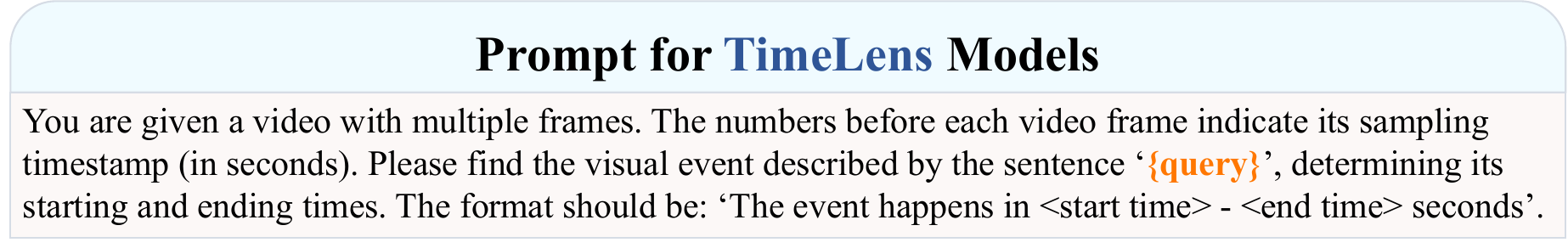}
   \caption{\textbf{Prompt for training and evaluating \paper.}
   }
   \label{fig:prompt_timelens}
\end{figure*}

\begin{figure*}[t]
  \centering
   \includegraphics[width=1\linewidth]{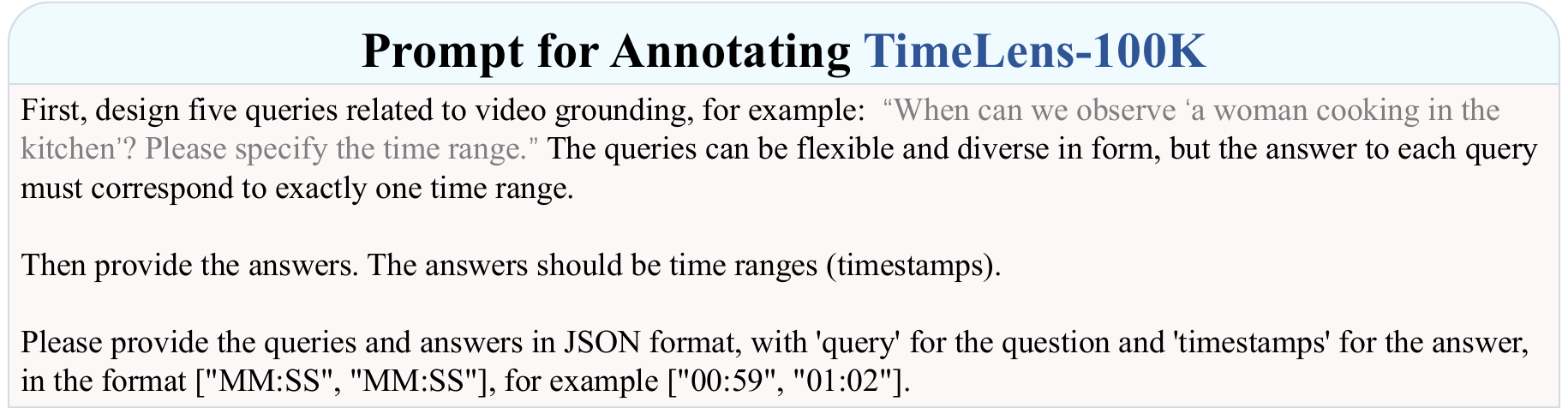}
   \caption{\textbf{Prompt for annotating \paper-100K.}
   }
   \label{fig:prompt_train_annotate}
\end{figure*}

\section{Details of Curating \paper-100K}
\label{sec:training_data_anno}

As described in~\cref{sec:training_data_anno}, we perform automated re-annotation on existing training datasets, resulting in \paper-100K, a large-scale, high-quality VTG training set comprising approximately 20K videos and 100K VTG annotations.

We begin by sampling videos from numerous existing VTG datasets, including CosMo-Cap~\cite{wang2024cosmo_cap}, InternVid-VTime~\cite{huang2024vtimellm-internvid-vtime}, DiDeMo~\cite{didemo}, QuerYD~\cite{oncescu2021queryd}, HiREST~\cite{zala2023hirest}, etc. These datasets already cover sufficiently diverse video domains. Additionally, we perform uniform sampling based on video duration to ensure sampled videos are approximately uniformly distributed within 0--240 seconds, with a small portion of longer videos included.

Given that most queries in the original annotations either lack clarity and specificity, or describe events that do not exist in the video, we directly use MLLMs for re-annotation. First, we prompt the MLLM to identify distinct events in the video and ensure these events are distributed across different time periods rather than being concentrated in a particular segment. Then, we have the model describe each event to generate queries and output the corresponding timestamps. Finally, we prompt the model to verify the quality of the queries and annotations.

Specifically, we use Gemini-2.5-Pro~\cite{comanici2025gemini-2-5}, currently the best-performing VTG model, for re-annotation. The annotation prompt is provided in~\cref{fig:prompt_train_annotate}. Notably, although this prompt appears simple, it is the result of extensive prompt engineering and optimization. We find that a concise and intuitive prompt is sufficient, as the model possesses adequate common sense and reasoning capabilities to understand the task. Overly complex and detailed prompts are unnecessary and can actually degrade annotation quality. During its reasoning process, the model can automatically verify and ensure the uniqueness and uniform distribution of events throughout the video.

As shown in~\cref{tab:train_data}, our training data substantially improves model performance, validating the enhanced quality of our training set. Notably, the construction of our training data is independent of our manual benchmark refinement process, ensuring a fair comparison.

\section{Implementation Details for Benchmarking Existing MLLMs}
In this section, we present the implementation details for evaluating existing MLLMs on our \paper evaluation suite, yielding the results reported in \cref{fig:performance_charades} and \cref{tab:main_result}.

\paragraph{\paper Models} The prompt for training and evaluating \paper models is illustrated in \cref{fig:prompt_timelens}. Implementation details are provided in \cref{sec:more_impl_details}.

\paragraph{GPT-5~\cite{OpenAI2025_GPT5} and GPT-4o~\cite{hurst2024gpt-4o}}
Since GPT models only support multi-image sequences as input, we sample frames from videos at 1 FPS and prepend textual timestamps (\textit{i.e.}, ``Frame at 2.5s:'') to each frame. As the Azure OpenAI API we use does not support more than 50 images for a single request, we adopt different strategies for videos longer than 50 seconds: for videos lasting 50-80 seconds, we uniformly sample 50 frames; for videos longer than 80 seconds, we sample at 1 FPS and arrange every 4 consecutive frames into a $2 \times 2$ grid within a single image, following previous works~\cite{ju2024miradata, zhang2024internlm-xcomposer-2-5, zhang2024longva}.
For GPT-5, which is a thinking model, we use the default value for the \texttt{reasoning.effort} parameter. The evaluation prompt is shown in \cref{fig:prompt_gpt}.

\paragraph{Gemini models~\cite{comanici2025gemini-2-5}}
Although Gemini models support audio input, we remove the audio from videos to ensure fair comparison with other vision-only models and maintain consistency with our benchmarks, which features vision-only, audio-free annotations. Following the best practices outlined in the official Gemini API documentation, we prompt the models to output timestamps in ``MM:SS'' format.
For other hyperparameters, we use the default settings: 1 FPS sampling and default \texttt{mediaResolution}, which tokenizes each frame into 258 tokens. For thinking models, we do not impose any limit on the \texttt{thinkingBudget} parameter. The evaluation prompt is shown in \cref{fig:prompt_gemini}.

\paragraph{Qwen3-VL~\cite{Qwen3-VL}, Qwen2.5-VL~\cite{qwen2-5-vl} and MiMo-VL~\cite{mimo-vl}}
\paper models, Qwen2.5-VL-7B, and MiMo-VL share approximately the same model architecture and hyperparameter configurations. Therefore, when evaluating these models, we adopt the same settings to ensure fair comparisons in~\cref{tab:main_result}. Specifically, consistent with~\cref{sec:exp_setup}, we sample video frames at 2 FPS and set the resolution budget to $\texttt{min\_tokens}=64$ and $\texttt{total\_tokens}=14,336$. For MiMo-VL, we evaluate their best-performing model, MiMo-VL-7B-RL. The evaluation prompt is shown in \cref{fig:prompt_qwen_mimo}.

\paragraph{Other Open-source Models} When evaluating Time-R1~\cite{wang2025timer1}, VideoChat-Flash~\cite{li2024videochat_flash} and VideoChat-R1~\cite{li2025videochat_r1}, we directly use their original codebases. Please refer to their papers and code repositories for details.

\begin{figure*}[t!] 
  \centering 

  \begin{minipage}{0.9\linewidth} 
    \centering 
    \includegraphics[width=\linewidth]{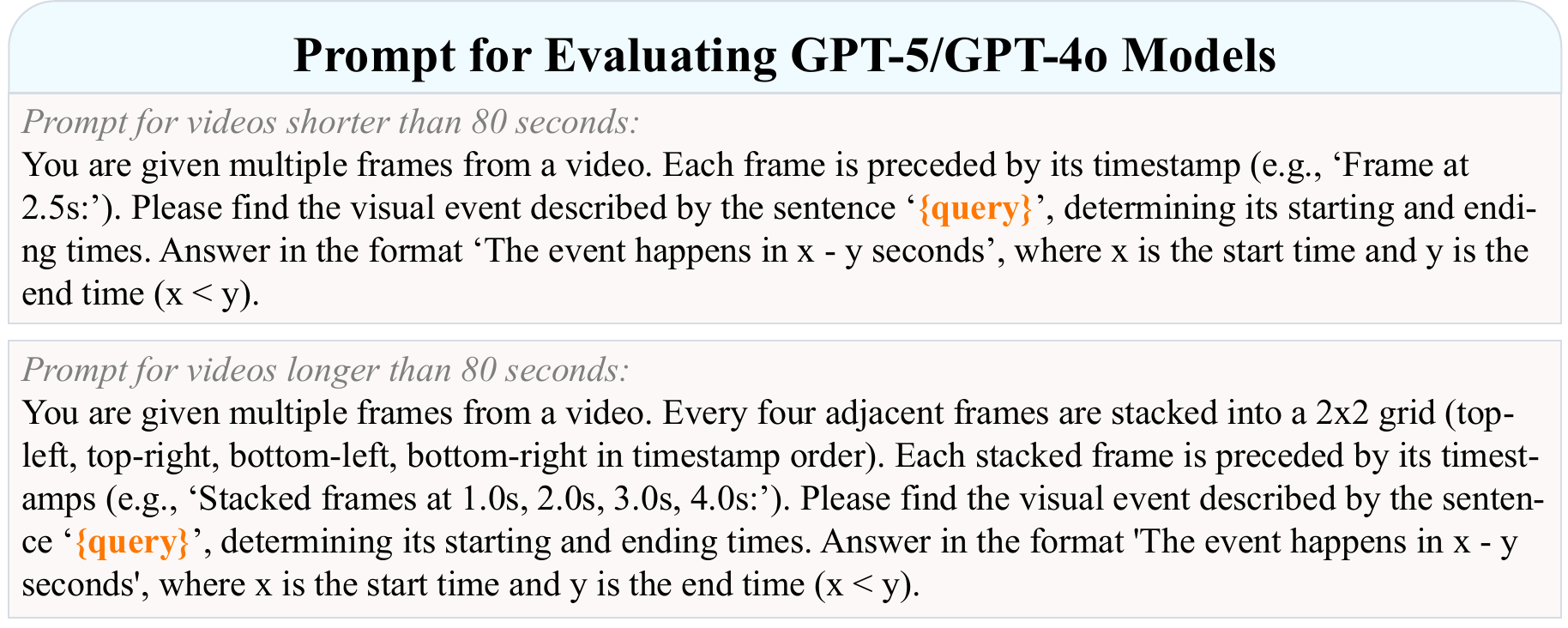}
      \vspace{-2em}
    \caption{\textbf{Prompts for evaluating GPT-5 and GPT-4o.}}
    \label{fig:prompt_gpt}
  \end{minipage}

  \vspace{1em} 

  \begin{minipage}{0.9\linewidth}
    \centering
    \includegraphics[width=\linewidth]{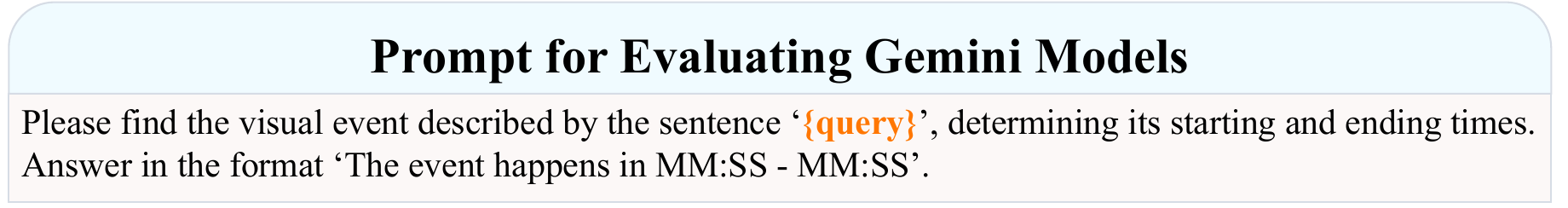}
  \vspace{-2em}
    \caption{\textbf{Prompt for evaluating Gemini models.}}
    \label{fig:prompt_gemini}
  \end{minipage}

  \vspace{1em}

  \begin{minipage}{0.9\linewidth}
    \centering
    \includegraphics[width=\linewidth]{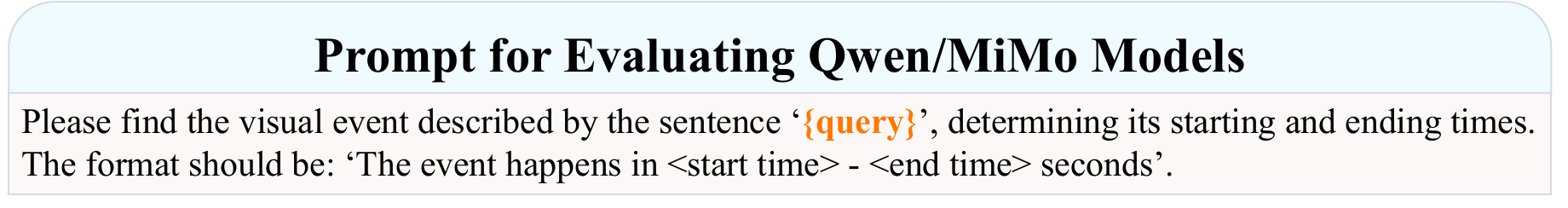}
          \vspace{-2.5em}
    \caption{\textbf{Prompt for evaluating Qwen3-VL, Qwen2.5-VL and MiMo-VL models.}}
    \label{fig:prompt_qwen_mimo}
  \end{minipage}

  \vspace{1em}

  \begin{minipage}{0.98\linewidth}
    \centering
    \includegraphics[width=\linewidth]{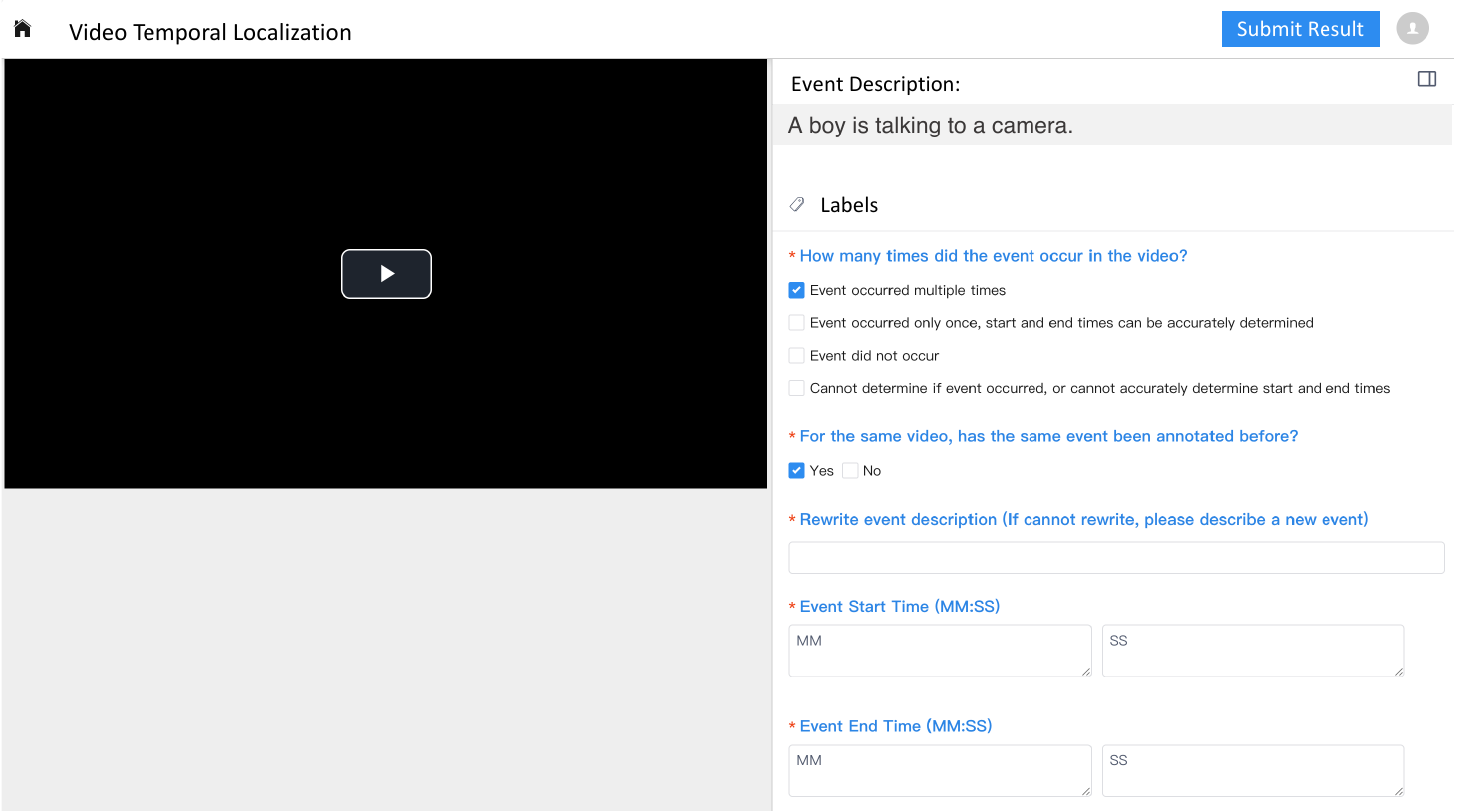}
    \caption{\textbf{Illustration of our annotation interface.}}
    \label{fig:anno_interface}
  \end{minipage}

\end{figure*}

\begin{figure*}[t]
  \centering
  \begin{subfigure}[b]{0.48\linewidth}
     \centering
     \includegraphics[page=1, width=\textwidth]{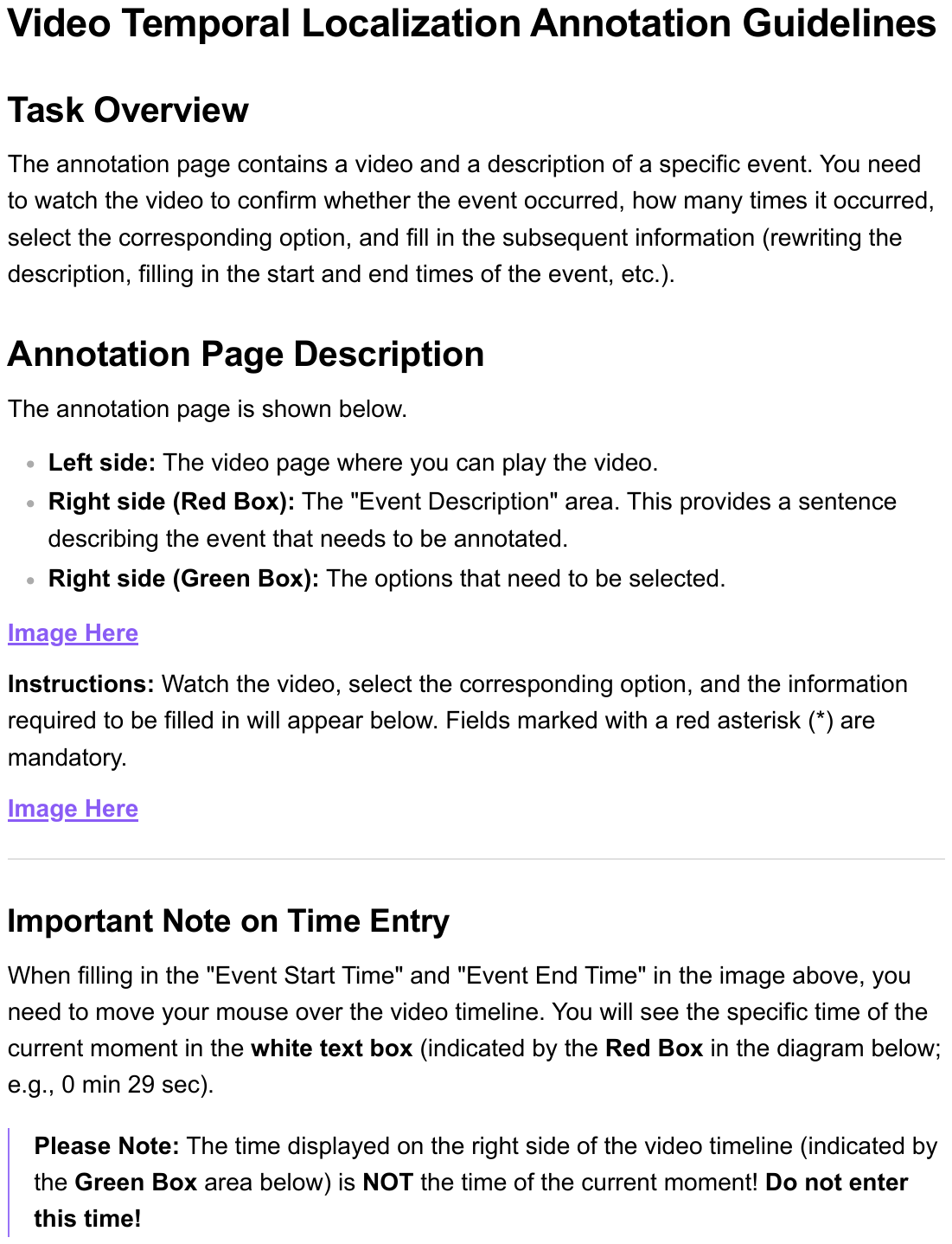}
          \vspace{-3em}
     \caption{}
  \end{subfigure}
  \hfill
  \begin{subfigure}[b]{0.48\linewidth}
     \centering
     \includegraphics[page=2, width=\textwidth]{assets/manual.pdf}
    \vspace{-3em}
      \caption{}
  \end{subfigure}

  \begin{subfigure}[b]{0.48\linewidth}
     \centering
     \includegraphics[page=3, width=\textwidth]{assets/manual.pdf}
     \vspace{-3em}
     \caption{}
  \end{subfigure}
  \hfill
  \begin{subfigure}[b]{0.48\linewidth}
     \centering
     \includegraphics[page=4, width=\textwidth]{assets/manual.pdf}
      \vspace{-3em}
    \caption{}
  \end{subfigure}

  \caption{\textbf{Illustration of our annotation manual.}
  Some figures and details are removed for confidentiality and safety reasons.}
  \label{fig:anno_manual}
\end{figure*}

\end{document}